\documentclass[journal]{IEEEtran}

\newcommand{\qed}{\nobreak \ifvmode \relax \else
      \ifdim\lastskip<1.5em \hskip-\lastskip
      \hskip1.5em plus0em minus0.5em \fi \nobreak
      \vrule height0.75em width0.5em depth0.25em\fi}

\ifCLASSINFOpdf
   \usepackage[pdftex]{graphicx}
\else
\fi
\usepackage{etoolbox}

\usepackage[cmex10]{amsmath}
\usepackage{amssymb}
\usepackage{notacion2}
\usepackage[normalem]{ulem}

\ifx\pdfoutput\undefined
\usepackage{graphicx}
\else
\usepackage[pdftex]{graphicx}
\usepackage{epstopdf}
\fi

\usepackage{environ}

\NewEnviron{equationw}{
\begin{equation}\begin{split}
\BODY
\end{split}\end{equation}
}

  \usepackage[caption=false,font=normalsize,labelfont=sf,textfont=sf]{subfig}

\usepackage{caption}
\usepackage{tikz}
\usepackage{textcomp}

\newcommand\copyrighttext{%
  \footnotesize \textcopyright 2018 IEEE. Personal use of this material is permitted.
  Permission from IEEE must be obtained for all other uses, in any current or future
  media, including reprinting/republishing this material for advertising or promotional
  purposes, creating new collective works, for resale or redistribution to servers or
  lists, or reuse of any copyrighted component of this work in other works.
  DOI: 10.1109/TNNLS.2018.2805019}
\newcommand\copyrightnotice{%
\begin{tikzpicture}[remember picture,overlay]
\node[anchor=south,yshift=5pt] at (current page.south) {\fbox{\parbox{\dimexpr\textwidth-\fboxsep-\fboxrule\relax}{\copyrighttext}}};
\end{tikzpicture}%
}

\newcommand\copyrighttextbis{%
  \footnotesize  This article has been accepted for publication in the journal IEEE Transactions on Neural Networks and Learning Systems, but has not been fully edited. Content may change prior to final publication. Citation information: DOI: 10.1109/TNNLS.2018.2805019, \textcopyright 2018 IEEE.}

\newcommand\copyrightnoticebis{%
\begin{tikzpicture}[remember picture,overlay]
\node[anchor=north,yshift=-2pt] at (current page.north) {\fbox{\parbox{\dimexpr\textwidth-\fboxsep-\fboxrule\relax}{\copyrighttextbis}}};
\end{tikzpicture}%
}

\begin{document}
%
\title{Complex Gaussian Processes for Regression}
%
%
%

\author{Rafael~Boloix-Tortosa*,
       Juan~Jos\'e~Murillo-Fuentes,~\IEEEmembership{Senior~Member,~IEEE,}
        F.~Javier~Pay\'an-Somet,
        and Fernando P\'erez-Cruz,~\IEEEmembership{Senior~Member,~IEEE,}
\thanks{R. Boloix-Tortosa, 
        Juan~Jos\'e~Murillo-Fuentes,
        and F.~Javier~Pay\'an-Somet are with the Departamento Teor\'ia de la Se\~nal y Comunicaciones, Escuela T\'ecnica Superior de Ingenier\'ia, Universidad de Sevilla, Camino de los Descubrimientos sn, 41092 Sevilla, Spain. e-mail: \{rboloix,murillo,jpayan\}@us.es. Fernando P\'erez-Cruz is with the Computer Science Department, Stevens Institute of Technology, 1 Castle Point Terrace,
Hoboken, NJ 07030 USA. e-mail: fperezcr@stevens.edu.}
\thanks{Thanks to Spanish government (Ministerio de Econom\'ia y Competitividad, TEC2016-78434-C3-02-R) and European Union (FEDER) for funding.
}
}

\hyphenation{op-tical net-works semi-conduc-tor hy-per-pa-ra-me-ters ge-ne-ra-tion ge-ne-ra-ted}

\onecolumn
\hspace{0pt}
\vfill
\begin{center}
\textbf{{\huge Complex Gaussian Processes for Regression}}
\\
\vspace{0.6cm}{\Large Rafael Boloix-Tortosa*, Juan Jos\'e Murillo-Fuentes*, F.~Javier~Pay\'an-Somet*, and Fernando P\'erez-Cruz$^\dagger$}
\\
\vspace{0.6cm}{\large * Dep. Signal Theory and Communications, Universidad de Sevilla, Spain.

$^\dagger$ Swiss Data Science Center, 8006 Zürich,
Switzerland and Department of Signal Theory and Communications, University Carlos III de Madrid, Spain.
}
\vspace{2cm}\\
\end{center}
{\Large This is the “accepted” version of the paper. Reference:}
\begin{table*}[ht]
\vspace{0.6cm}
\centering
\begin{tabular}{l}
{\Large Final Title: "Complex Gaussian Processes for Regression"} \\[10pt]
{\Large Journal: IEEE Transactions on Neural Networks and Learning Systems}\\[10pt]
{\Large DOI: 10.1109/TNNLS.2018.2805019} \\[10pt]
\end{tabular}
\end{table*}
\vspace{2cm}

{\large{\textcopyright 2018 IEEE. Personal use of this material is permitted. Permission from IEEE must be obtained for all other uses, in any current or future media, including reprinting/republishing this material for advertising or promotional purposes, creating new collective works, for resale or redistribution to servers or lists, or reuse of any copyrighted component of this work in other works.}}
\vfill
\hspace{0pt}

\twocolumn
\newpage

\maketitle

\copyrightnoticebis
\copyrightnotice

\begin{abstract}

In this paper we propose a novel Bayesian solution for nonlinear regression in complex fields. Previous solutions for kernels methods usually assume a \emph{complexification} approach, where the real-valued kernel is replaced by a complex-valued one. This approach is limited. Based on results in complex-valued linear theory and Gaussian random processes we show that a \emph{pseudo-kernel} must be included.
This is the starting point to develop the new complex-valued formulation for  Gaussian process for regression (CGPR). 
We face the design of the covariance and pseudo-covariance based on a convolution approach and for several scenarios. Just in the particular case where the outputs are proper, the pseudo-kernel cancels.  Also, the hyperparameters of the covariance {can be learnt} maximizing the marginal likelihood using Wirtinger's calculus and patterned complex-valued matrix derivatives.  In the experiments included, we show how CGPR successfully solve systems where real and imaginary parts are correlated. Besides, we successfully solve the nonlinear channel equalization problem by developing a recursive solution with basis removal. We report remarkable improvements compared to previous solutions: a 2-4 dB reduction of the MSE with {just a quarter} of the training samples used by previous approaches.

 \end{abstract}

\begin{IEEEkeywords}
Gaussian processes, regression, complex-valued processes, kernel methods.
\end{IEEEkeywords}

%

\setlength{\arraycolsep}{0.1em} 

\section{Introduction}
%
%
%
%

\IEEEPARstart{N}{owadays} complex-valued signals model a vast range of nowadays systems in science and engineering such as telecommunications, optics {and acoustics among others. Complex-valued signal processing allows to natively process complex-valued sequences, like electromagnetic signals}. {Hence}, complex-valued signal processing is of fundamental interest. 
Signal processing for complex-valued signals has been widely studied in the linear case, see \cite{Schreier06} and references therein. 
The nonlinear processing of complex-valued signals has been addressed from \notaRBT{different points of view, such as complex-valued nonlinear adaptive filtering \cite{Mandic09},} neural networks \cite{hirose13}, \notaRBT{\cite{Valle14}} and, recently, using reproducing kernel Hilbert spaces (RKHS) \cite{Scholkopf02}. Some complex kernel-based algorithms have been lately proposed for classification \cite{Steinwart06}, regression \cite{OgunfunmiP11,Bouboulis12,Tobar12} and mainly for kernel principal component analysis \cite{Papaioannou14}. 
Regarding regression, in \cite{Bouboulis11} the authors propose a complex-valued kernel {based on} the results in \cite{Steinwart06} and face the derivative of cost functions by using Wirtinger's derivatives. Same kernel \notaRBT{is} adopted in \cite{OgunfunmiP11}. 
In \cite{Tobar12} the authors review the kernel design to improve the previous solutions. 
These previous approaches have been developed in the framework of kernel least-mean-square {(KLMS)}. In the framework of kriging some complex-valued scenarios have also been addressed \cite{Lajaunie91,DeIaco03,DeIaco16}.
%

The methods above proposed for regression deal with complex-valued inputs and outputs by {either} 1) learning the real and imaginary parts independently{;} 2) using {a} straightforward adaptation of a real-valued approach{;} or{,} 3) learning a vector with the real and imaginary parts stacked {in an augmented vector}. The first approach is suboptimal for systems where the real and imaginary parts are not independent. {T}he straightforward adaptation of real-valued versions is limited to proper systems{,} as \emph{strictly} linear approaches \cite{Schreier06}. \emph{Complexification} of real RKHSs \cite{Paulsen09} lies within this group. For non-proper systems \emph{widely} linear solutions are needed. 
The last {option} fits any scenario, but the complex valued formulation is lost, {which limits the native interpretation of the complex sequence and applicability of this procedure. Furthermore,} the design of the kernel between the real and the imaginary parts remains an open problem. To the best of our knowledge, except for \cite{Bouboulis12}, where an augmented version is discussed for the KLMS, there is no general complex-valued formulation of a nonlinear regression algorithm based on kernels or covariances. In this paper, we propose a new complex-valued algorithm working both for proper and non-proper systems.

\notaRBT{The Gaussian process (GP) \cite{Rasmussen06} is a Bayesian nonparametric framework for inference. It has has attracted increasing attention from the machine learning community for its many applications, such as in regression, classification \cite{Muller01,Li17}, adaptive control \cite{Chowdhary15}, multitask learning \cite{Skolidis13}, or data association \cite{Lazaro14}, among others.} 
{Gaussian processes for regression (GPR) \cite{Rasmussen06}, \notaRBT{\cite{Grande17}}, \cite{PerezCruz13gp} are kernel methods that provide a full conditional statistical description for the predicted variable. The covariance matrix of the GPR plays the role of the kernel. 
In \cite{boloix14} we developed complex-valued GPR for proper systems. A proper complex random signal is uncorrelated with its complex conjugate \cite{Neeser93}, and hence the pseudo-covariance cancels. The solution in \cite{boloix14} can be described as \notaRBT{a} straightforward adaptation of a real-valued approach as in \cite{Scholkopf02,OgunfunmiP11,Bouboulis12,Tobar12}. In this paper we include the pseudo-covariance of a Gaussian process into GPR to develop a novel approach for complex-valued systems, hereafter denoted as complex GPR (CGPR). With this result we prove that another kernel matrix is needed to properly model any given system, including all non-proper ones. We also tackle the maximization of the marginal likelihood in the complex-valued scenario. Since it depends on a complex Hermitian matrix, generalized complex-valued derivatives are used \cite{Hjorungnes08}.  }

The design of a good covariance and pseudo-covariance function is crucial for CGPRs to provide accurate nonlinear solutions. Under the Gaussian process regression point of view, the covariance function measures \emph{similarity} between inputs \cite{Rasmussen06}. 
The construction of the imaginary part quite depends on the system model. We propose to apply the convolution approach \cite{Boyle05,Calder07} to generate a covariance and pseudo-covariance that explain the case where real and imaginary parts are correlated, even when shifted or displaced. 
This {includes} the proper case where the cross covariance between the real and the imaginary parts is {either null or} skew-symmetric.

As benchmark we propose the nonlinear channel equalization problem in digital communications. The authors in \cite{Bouboulis11,Bouboulis12} propose to build KLMS adaptive filtering of complex signals.
We face this problem from the CGPR point of view. The statistical properties of the to-be-learned outputs in the channel equalization problem are taken into consideration in the selection of the model and a recursive version with basis removal criterion is adapted \cite{Vaerenbergh12}. Compared to the solutions in \cite{Bouboulis11,Bouboulis12,Tobar12}  the CGPR exhibits a {2-4 dB gain with only a quarter} of the training samples used by {state-of-the-art} approaches. Other experiments are also included to analyze other scenarios.

The paper is organized as follows. Next section includes the definition of the CGPR, including the derivation of the proper case for CGPR in \SSEC{PCGPR}. 
\SEC{Cov} is devoted to the analysis of the structure of the complex-valued covariance and pseudo-covariance. We develop in \SEC{ML} the optimization procedure to set the kernel hyperparameters applying Wirtinger's calculus and patterned complex-valued matrix derivatives. {We show the experimental results in Section V and conclude the paper in Section VI.}

The notations used in the paper are as follows. For matrix $\vect{A}$, $\det\vect{A}$ is its determinant, {$\mathrm{Tr}(\vect{A}$) is its trace}, $\entry{\vect{A}}{l}{q}$ is its $(l,q)$ entry, $\vect{A}\trs$ represents the transpose, $\vect{A}\her$ the Hermitian transpose, $\vect{A}^{*}$ represents the complex conjugate of its entries, and $\matr{A}^{-*}=(\vect{A}^{*})\inv$. 
To denote the $i$-th sample of a vector we use $\vect{a}_i$. The real and imaginary parts are denoted by subindices $\textrm{r}$ and $\textrm{j}$, respectively, i.e. $\vect{a}=\vect{a}\rr+\j \vect{a}\jj$, with $\textrm{j}=\sqrt{-1}$. \notaRBT{$\mathbb{E}[\cdot]$ refers to expectation}. To denote the complex Gaussian distribution with mean vector $\boldsymbol{\mu}$, covariance matrix $\K$ and pseudo-covariance matrix $\matr{\tilde{K}}$ we use $\calg{N}\left(\boldsymbol{\mu},\K,\matr{\tilde{K}}\right)$. The augmented covariance matrix, $\aug{\K}$, is given by
\begin{equation}\LABEQ{augK}
\aug{\K}=\left[\begin{array}{lll}
\K & {} & \psc{\K}\\
\psc{\K}^* & {} & \K^*\\
\end{array}\right].
\end{equation}


\section{Complex Gaussian Process Regression}\LABSEC{CGPR}
\subsection{Complex-valued Gaussian process}
GPR can be presented as a nonlinear regressor that expresses the input-output relation through function $f(\x)$, known as latent function. This latent function follows a GP and underlies the regression problem
\begin{equation}\LABEQ{regression}
{\y}=\f(\x)+\epsilon,
\end{equation}
where the error, $\epsilon$, in the estimation of a real-valued output, $\y$, is modeled as additive zero-mean Gaussian noise.
In this paper, we consider that both inputs and outputs are complex-valued. The simpler real-valued input and complex-valued output case can be easily solved from the results herein. Each input at time $i$ is a complex-valued column vector of dimension $\d$, $\x_i \in {\CN}^d$. For any input set {$\X\subn=[\x_1, . . . , \x_\n]$} the latent function in \EQ{regression} provides a multidimensional Gaussian complex-valued random vector $\fv\subn=[f(\x_{1}), . . . , {f}(\x_\n)]^{\top}$, where $f(\x_i)\in \CN$.  
A complex random Gaussian vector $\fv\subn$ is characterized not only by its mean vector $\boldsymbol{\mu}$ and covariance matrix $\K= \mathbb{E}[(\fv\subn-\boldsymbol{\mu})(\fv\subn-\boldsymbol{\mu})\her]$, but also by its complementary covariance or pseudo-covariance matrix $\psc{\K}=\mathbb{E}[(\fv\subn-\boldsymbol{\mu})(\fv\subn-\boldsymbol{\mu})^\top]$, \cite{Schreier06}. These matrices can be defined by kernels, $\entry{\vect{\K}}{l}{q}=\k(\x_l,\x_q)$
 and $\entry{\vect{\psc{\K}}}{l}{q}=\psc{\k}(\x_l,\x_q)$, respectively. The Gaussian process prior becomes
\begin{align}\LABEQ{complexgaussian}
&p(\fv\subn|\X\subn)=\calg{N}\left(\boldsymbol{\mu},\K,\matr{\tilde{K}}\right)\nonumber\\
&=\frac{1}{\pi^n \sqrt{\det\aug{\K}}}\exp{\left(-\frac{1}{2}(\aug{\fv\subn}-\aug{\boldsymbol{\mu}})\her\aug{\K}\inv(\aug{\fv\subn}-\aug{\boldsymbol{\mu}})\right)},
\end{align}
where $\aug{\fv\subn}=[{\fv\subn}^\top \;{\fv\subn}\her]^\top$ is the augmented vector for $\fv\subn$, $\aug{\boldsymbol{\mu}}=[\boldsymbol{\mu}^\top \;\boldsymbol{\mu}\her]^\top$ is the augmented mean vector, and $\aug{\K}$ is the augmented covariance matrix \EQ{augK}. 
{Without loss of generality, we} consider zero-mean processes, $\mu(\x)=0$. 

\subsection{Complex GP for Regression} 
In the learning process we condition the output of the GPR for some new observation 
 ${\x\new}$,
given the training set $\tset\subn=\{\X\subn,\yv\subn\}$, where the outputs $\yv\subn=\left[{\y_1}, . . . , {\y_\n}\right]^{\top}$ for a given set of observations $\X\subn$ are known. First, we compute the joint distribution {as} follows. We assume that the additive noise $\epsilon$ in \EQ{regression} follows an i.i.d. {complex} Gaussian distribution with zero mean, variance $\sigma^2_\epsilon$ and {pseudo-covariance} $\rho\sigma_{\epsilon}^2$, \notaRBT{with $\rho$ being} a complex number.
The samples in the training set are i.i.d., hence the likelihood for the latent function at the training set is given by the factorized model
\begin{equation}\LABEQ{factmod}
p\left(\yv\subn| \fv\subn \right)= \prod_{i=1}^n {p(y_i|f(\x_i))},
\end{equation}
where $p(y_i|f(\x_{i}))=\calg{N}(f(\x_{i}),\sigma^2_\epsilon,\rho\sigma_{\epsilon}^2)$\footnote{If the likelihood were not Gaussian, we can resort to Wrapped Gaussian processes \cite{Snelson04, Lazaro12}.}.
Therefore, the likelihood is a complex multidimensional Gaussian $p(\yv_{\n}| \fv\subn )= \calg{N}(\fv\subn,\sigma^2_\epsilon\id{\n},\rho\sigma^2_\epsilon\id{\n})$. This likelihood and the prior in \EQ{complexgaussian} yield the marginal likelihood or evidence 
\begin{equation}\LABEQ{marlikelihood}
p\left(\yv\subn| \X \right)=\int{p\left(\yv\subn| \fv\subn \right)p(\fv\subn|\X\subn)d\fv\subn} =\calg{N}\left(\vect{0},\C,\psc\C\right),
\end{equation}
where  $\C=\K+\sigma_\epsilon^2\id{\n}$ and $\psc\C=\psc{\K}+\rho\sigma^2_\epsilon\id{\n}$.
Given a test input vector $\x\new$, the joint distribution of the training outputs $\yv\subn$ and ${\f\new}={f}(\x\new_1)$ is
\begin{equation}\label{eq:yf}
\left[ \begin{array}{c}
\yv\subn\\
{\f\new}\\
\end{array}\right]\sim\calg{N}\left( \matr{0}, \matr{\Lambda},\psc{\matr{\Lambda}}\right),
\end{equation}
with
\begin{align}\label{eq:Lambda}
\matr{\Lambda}&=\left[ \begin{array}{cc}
\C & \kv(\X\subn,\x\new)\\
\kv\her(\X\subn,\x\new) &\k(\x\new,\x\new)\\
\end{array}\right], \\
\psc{\matr{\Lambda}}&=\left[ \begin{array}{ccc}
\psc\C & \psc\kv(\X\subn,\x\new)\\
\psc\kv\trs(\X\subn,\x\new) & \psc\k(\x\new,\x\new)\\
\end{array}\right]
\end{align}
where $\kv(\X\subn,\x\new)=[\k(\x({1}),\x\new),\cdots,\k(\x({n}),\x\new)]\trs$ and $\psc\kv(\X\subn,\x\new)=[\psc\k(\x({1}),\x\new),\cdots,\psc\k(\x({n}),\x\new)]\trs$. 

The conditional distribution of ${\f\new}$ given $\yv\subn$, i.e. the estimated output, yields the Gaussian distribution
$p\left({\f\new}| {\x\new},\X\subn,\yv\subn\right)=\calg{N}({\mu}_{{\f\new}},{\sigma}_{\f\new},\psc{\matr{\sigma}}_{\f\new}), 
$ {where}
\begin{align}
\aug{\vect{\mu}}_{\f\new}&=\left[ \begin{array}{c}
{{\mu}}_{\f\new}\\
{{\mu}}_{\f\new}^{*}\\
\end{array}\right]=\aug{\K}\her(\X\subn,\x\new)\aug{\matr{C}}\inv\aug{\yv}\subn,\label{eq:meancomplex}\\
\aug{\matr{\Sigma}}_{\f\new}&=\left[ \begin{array}{cc}
{\sigma}_{\f\new} & \psc{\sigma}_{\f\new}\\
\psc{\sigma}\cnj_{\f\new} & {\sigma}_{\f\new}\\
\end{array}\right]=\aug{\K}(\x\new,\x\new)-\aug{\K}\her(\X\subn,\x\new)\aug{\matr{C}}\inv\aug{\K}(\X\subn,\x\new),\label{eq:varcomplex}
\end{align}
{and}
\begin{align} 
\aug{\K}(\x\new,\x\new)&=\left[ \begin{array}{lll}
\k(\x\new,\x\new)&\psc\k(\x\new,\x\new)\\
\psc\k\cnj(\x\new,\x\new)&\k\cnj(\x\new,\x\new)\\\end{array}\right],\\
\aug{\K}(\X\subn,\x\new)&=\left[ \begin{array}{lll}
 \kv(\X\subn,\x\new)& {} & \psc\kv(\X\subn,\x\new)\\
\psc\kv\cnj(\X\subn,\x\new)& {} & \kv\cnj(\X\subn,\x\new)\\
\end{array}\right].
\end{align}
$\aug{\C}$ is the augmented covariance matrix of the augmented observations $\aug{\yv}\subn=[{\yv\subn}^\top \;{\yv\subn}\her]^\top$.
By using the matrix-inversion lemma 
\begin{align}
\aug{\C}\inv=\left[\begin{array}{lll}
\C & {} & \psc{\C}\\
\psc{\C}^* & {} & \C^*\\
\end{array}\right]\inv=\left[ \begin{array}{cc}
\matr{P}\inv & -\C\inv\psc\C\matr{P}^{-*}\\
-\C^{-*}\psc\C^{*}\matr{P}\inv & \matr{P}^{-*}\\
\end{array}\right]\hspace{-2 pt},
\end{align}
where $\matr{P}=\C-\psc\C\C^{-*}\psc\C^{*}$. Therefore, the mean, covariance and pseudo-covariance of the prediction yield, respectively, 
\begin{align}\label{eq:meanfnew}
\vect{\mu}_{{\f}\new}&=\left[\kv\her(\X\subn,\x\new)- \psc\kv\trs(\X\subn,\x\new)\C^{-*}\psc\C^{*}\right]\matr{P}\inv\yv\subn\nonumber\\
&+\left[ \psc\kv\trs(\X\subn,\x\new)-\kv\her(\X\subn,\x\new)\C\inv\psc\C\right]\matr{P}^{-*}\yv\subn^{*},\\
\label{eq:varfnew}
{\sigma}_{\f\new} &=\k(\x\new,\x\new)\nonumber\\&-\left[\kv\her(\X\subn,\x\new)- \psc\kv\trs(\X\subn,\x\new)\C^{-*}\psc\C^{*}\right]\matr{P}\inv\kv(\X\subn,\x\new)\nonumber\\
&-\left[ \psc\kv\trs(\X\subn,\x\new)-\kv\her(\X\subn,\x\new)\C\inv\psc\C\right]\matr{P}^{-*}\psc\kv^{*}(\X\subn,\x\new),\\
\label{eq:pvarfnew}
\psc{\sigma}_{\f\new}&=\psc\k(\x\new,\x\new)\nonumber\\&-\left[\kv\her(\X\subn,\x\new)- \psc\kv\trs(\X\subn,\x\new)\C^{-*}\psc\C^{*}\right]\matr{P}\inv\psc\kv(\X\subn,\x\new)\nonumber\\
&-\left[ \psc\kv\trs(\X\subn,\x\new)-\kv\her(\X\subn,\x\new)\C\inv\psc\C\right]\matr{P}^{-*}\kv^{*}(\X\subn,\x\new).
\end{align}

\subsection{Proper Complex GPR}\LABSSEC{PCGPR}

When the pseudo-covariance matrix cancels, a complex Gaussian random vector is regarded as \emph{proper} \cite{Neeser93,Ollila08}. 
In the zero-mean proper case, the prior \EQ{complexgaussian} simplifies to
\begin{equationw}\label{eq:propergaussian}
p(\fv\subn|\X\subn)=\calg{N}\left(\vect{0},\K,\matr{0}\right)=\frac{1}{\pi^n \det{\K}}\exp{\left(-{\fv\subn}\her{\K}\inv{\fv\subn}\right)},
\end{equationw}
and the marginal likelihood \EQ{marlikelihood} to $p(\yv\subn| \X\subn)=\calg{N}(\vect{0},\C,\matr{0})$, so that $\yv\subn$ is also proper Gaussian. Furthermore, $\yv\subn$ and $\fv\subn$ are cross-proper, {i.e.,} the complementary cross-covariance matrix $\mathbb{E}[\yv\subn\fv\subn^\top]=\matr{0}$.
Hence, $\yv\subn$ and $\fv\subn$ are jointly proper \cite{Schreier06}, i.e., the composite complex random vector $[\yv\subn^\top,\fv\subn^\top]^\top$ is proper Gaussian. Now, given a test input vector $\x\new$, the joint distribution of the training outputs $\yv\subn$ and ${\f\new}$ is: 
\begin{equation}\label{eq:yfp}
\left[ \begin{array}{c}
\yv\subn\\
\f\new\\
\end{array}\right]\sim\calg{N}\left( \matr{0}, \matr{\Lambda},\matr{0}\right).
\end{equation}
{The} estimated probabilistic output is the conditional distribution of ${\f\new}$ given $\yv\subn$:
\begin{equation} \label{eq:posteriorfprop}
p\left({\f\new}| {\x\new},\X,\yv\subn\right)=\calg{N}\left({\mu}_{{\f\new}},{\sigma}_{\f\new},\matr{0}\right),
\end{equation}
which is the conditional proper complex Gaussian distribution
described with the following mean vector and covariance matrix 
\begin{align}
\vect{\mu}_{{\f}\new}&=\kv\her(\X\subn,\x\new)\C\inv\yv\subn, \LABEQ{mediaCPGP}\\
{\sigma}_{\f\new} &=\k(\x\new,\x\new)-\kv\her(\X\subn,\x\new)\C\inv\kv(\X\subn,\x\new).\label{eq:varCPGP}
\end{align}
{Notice that this result is the straightforward adaptation of the real-valued GPR to complex-valued signals, where transpose is changed to Hermitan transpose.} 
%

{\section{Complex covariance functions}\LABSEC{Cov}}

Under the GPR point of view, the covariance function should measure \emph{similarity} between inputs \cite{Rasmussen06,PerezCruz13gp}. A usual option is to consider that training points that are near to a test point are informative about the prediction at that point. In other kernels, e.g., polynomial kernels, similarity is measured in a different way. Covariance matrices should also be semi-definite positive. We next develop these issues for the complex-valued case, where we have the covariance and pseudo-covariance matrices. 
%
Given a zero-mean complex Gaussian vector $\fv\subn={\fv\subn}\rr+\j{\fv\subn}\jj$, with ${\fv\subn}\rr$ its real part and ${\fv\subn}\jj$ its imaginary part: 
\begin{align}
\K&=\mathbb{E}\left[\fv\subn\fv\subn\her\right]=\K\rrrr+\K\jjjj+\j\left(\K\jjrr-\K\rrjj\right),\LABEQ{covK}\\
\psc\K&=\mathbb{E}\left[\fv\subn\fv\subn^\top\right]=\K\rrrr- \K\jjjj+\j\left(\K\jjrr+\K\rrjj\right),\LABEQ{pcovK}
\end{align}
where $\K\rrrr=\mathbb{E}[{\fv\subn}\rr{\fv\subn}\rr\trs]\in\RN_+^{n\times n}$ and $\K\jjjj=\mathbb{E}[{\fv\subn}\jj{\fv\subn}\jj\trs]\in\RN_+^{n\times n}$ are the covariance matrices of the real and imaginary parts of $\fv\subn$, respectively, and $\K\rrjj= \mathbb{E}[{\fv\subn}\rr{\fv\subn}\jj^\top]=\K\jjrr\trs$ $\in\RN^{n\times n}$ is the cross-covariance matrix of the real and imaginary parts. 
Matrix ${\K}$ must be Hermitian positive semidefinite while $\psc\K$ must be symmetric. From the augmented point of view, the Schur complement of the augmented covariance matrix $\aug{\K}$ must be positive semidefinite \cite{Schreier06}. 

In the design of these matrices we may proceed as follows.
On the one hand, we can directly construct complex-valued functions that produce matrices as \EQ{covK} and \EQ{pcovK} with the properties described above, i.e., ${\K}$ must be Hermitian positive semidefinite and $\psc\K$ must be symmetric. Those complex-valued functions should be carefully selected in order to fairly represent the covariance and pseudo-covariance properties of the complex-valued process being modeled. On the other hand, we may try to design their real and imaginary parts. In this second method, we can resort to three real functions $\k\rrrr(\x,\x')$, $\k\jjjj(\x,\x')$ and $\k\rrjj(\x',\x)$ of the complex inputs $\x$, that are used to write out the three real matrices $\K\rrrr$, $\K\jjjj$ and $\K\jjrr=\K\rrjj\trs$. Again, the resulting covariance matrix ${\K}$ must be Hermitian positive semidefinite and $\psc\K$ must be symmetric, and meeting these conditions from the design of their parts is not straightforward. However, this second option provides one important advantage:
known correlation properties of the real and imaginary parts can be translated directly into the covariance and pseudo-covariance functions. 

One important example is when it is known that the real and imaginary parts are uncorrelated and have null cross-covariance matrix. In such a case we should set $\k\rrjj(\x',\x)=0$ and the covariance and pseudo-covariance functions yield real functions. Also, any information about stationarity, periodicity, etc. of the real part can be modeled in $\k\rrrr(\x,\x')$, and the same can be said about the imaginary part and $\k\jjjj(\x,\x')$.
Furthermore, in the particular case of a proper complex Gaussian vector the pseudo-covariance matrix \EQ{pcovK} nulls and we can resort to \emph{proper} complex GPR. Hence, $\K\rrrr=\K\jjjj$ and $\K\jjrr=\K\rrjj\trs=-\K\rrjj$, i.e., $\K\rrjj$ is a skew-symmetric cross-covariance matrix. In this case, the covariance matrix \EQ{covK} simplifies to $\K=2\K\rrrr-2\j\K\rrjj$, and the following properties for the three proposed real functions hold: $k\rrrr(\x,\x')=k\jjjj(\x,\x')$ and $k\rrjj(\x,\x')=-k\rrjj(\x',\x)$. Also, as the covariance matrix  must be Hermitian positive semi-definite, it follows that $k\rrrr(\x,\x')$ and $k\rrjj(\x',\x)$ are interrelated. 

Finally, the way that similarity is measured in the covariance and pseudo-covariance functions is another important issue to take into account when selecting them for complex-valued GPR {and the similarity must be measured in the complex field}. 

\subsection{Examples of complex-valued kernels and covariances functions}

We first recall some examples of complex-valued kernels functions found in the literature.

\subsubsection{Complex-valued Gaussian kernel} 
The first example is the complex-valued Gaussian kernel \cite{Steinwart06,Bouboulis11}. It is  an extension of the real Gaussian kernel defined as 
\begin{align}\LABEQ{RK}
\k_{\CN}(\x,\x')=\exp\left(-({\x-\x'^{*})}^\top(\x-\x'^{*})/\gamma\right),
\end{align}
{with kernel {hyperparameter} $\gamma$}.
If we separate the real and imaginary parts of the kernel
\begin{align}\LABEQ{CGK}
\k_{\CN}(\x,\x')
&=\exp\left(-|\x\rr-\x\rr'|^2/\gamma\right)\exp\left(|\x\jj+\x\jj'|^2/\gamma\right)\nonumber\\
&\cdot\left(\cos(2(\x\rr-\x\rr')\trs(\x\jj+\x\jj')/\gamma)\right.\nonumber\\&\left.-\j\sin(2(\x\rr-\x\rr')\trs(\x\jj+\x\jj')/\gamma)\right),
\end{align}
where $|\cdot|$ is the $\ell^2$-norm. Note that this kernel gives rise to a covariance matrix with skew-symmetric cross-covariance matrix $\K\rrjj$. Hence, it fits in the proper case with a null pseudo-covariance.
%
This kernel does not provide its maximum when $\x=\x'$, but when $\x=\x'^*$, i.e., it measures similarities between the real parts of the inputs, while measures dissimilarity between imaginary ones. {The} value it provides when $\x=\x'$ is not constant but depends on the imaginary part of $\x$ as $\exp(|2\x\jj|^2/\gamma)$. Also, it is not stationary, it has an oscillatory behavior and may also cause serious numerical problems in the learning algorithms, as is later discussed in the Experiments section. 

\subsubsection{Independent kernel}
In \cite{Tobar12} the following kernel was proposed:
\begin{align}\LABEQ{indepCGK}
\k_{ind}(\x,\x')&=\kappa_{\Rext}\left(\x\rr,\x'\rr\right)+\kappa_{\Rext}\left(\x\jj,\x'\jj\right)\nonumber\\
&+\j\left(\kappa_{\Rext}\left(\x\rr,\x'\jj\right)-\kappa_{\Rext}\left(\x\jj,\x'\rr\right)\right),
\end{align}
where $\kappa_{\Rext}$ is a real kernel of real inputs, in particular, they propose the real Gaussian kernel. Notice that this is an example of a design using real-valued functions, $\k\rrrr(\x,\x')$, $\k\jjjj(\x,\x')$ and $\k\rrjj(\x',\x)$, but with two simplifications. First, the three functions are the same real function $\kappa_{\Rext}$. Second, the inputs of the function are not complex, but real, i.e., the real part or the imaginary part of $\x$. Because of this simplifications, the independent kernel provides a high value when the inputs have equal real parts, $\x\rr=\x'\rr$, although the imaginary parts are very different. We have the same behavior for the imaginary part. 
Also, note that this kernel gives rise to a covariance matrix with skew-symmetric cross-covariance matrix $\K\rrjj$. Hence, it also could be used in the proper case as covariance.

\subsubsection{Spectral kernels}
Finally, there have also been some proposals to create an imaginary part from the real part of the covariance function in kriging \cite{Lajaunie91,DeIaco03}  for a multiple output learning framework. {However, these proposals are for stationary random fields of {just real} inputs, i.e., $k(\x-\x')$, with $\x\in\RR^d$, and do not provide a pseudo-covariance {function.} 
In \cite{Lajaunie91}, they propose to obtain a covariance matrix starting from a given function. However, it is unclear what the function should be for the covariance matrix to have some given properties. In \cite{DeIaco03} the proposed covariance function exhibits a sinusoidal behavior in its real and imaginary parts that is in general not suitable for the application at hand.}

\subsection{Convolution approach} \LABSSEC{convmeth}

We propose {to follow} the convolutional approach \cite{Boyle05,Calder07} as a link between the two points of view in the design of the covariance functions. The idea is to generate a complex random process as the sum of the outputs of linear filters driven by real white noises. The starting point is the selection of functions that provide the desired measures of similarity and modeling for the covariances. Those functions are used as filters to generate a random process. The calculation of the covariance and pseudo-covariance of the process {provides} the covariance and pseudo-covariance functions. This way we ensure that the resulting covariance and pseudo-covariance functions {are} valid, i.e., {generat{e}} a valid hermitian ${\K}$ and a valid symmetric $\psc\K$. 
Then, the design of the filters conditions the properties of the kernels and the associated similarity between pairs of inputs. 

Let $U(\x)$ be the complex process written as the output of linear filters:
\begin{align} \LABEQ{U}
U(\x)=&
\left(h_{1}(\x)+\j h_{2}(\x)\right) \star S\rr(\x)\nonumber\\&
+\left(h_{3}(\x)+\j h_{4}(\x)\right) \star S\jj(\x)
\end{align}
where {$h_m(\x),\,m\in\{1,2,3,4\}$ represent the filters and} $\star$ denotes the convolution operation. $S\rr(\x)$ and $S\jj(\x)$ are {independent} real white noises with zero mean and unit variance. The inputs are complex-valued vectors $\x\in\CC^\d$. 
The covariance of $U(\x)$ for two different inputs is used as covariance function, $\k(\x,\x')=\C_{U}(\x,\x')=\E\left[U(\x)U^*(\x')\right]$, and the pseudo-covariance is used as pseudo-covariance function, $\psc\k(\x,\x')=\psc\C_{U}(\x,\x')=\E\left[U(\x)U(\x')\right]$. Details of the calculations of $\C_{U}(\x,\x')$ and $\psc\C_{U}(\x,\x')$ can be found in \APEN{Apen1}.
%
The complexity here lies in the selection of the filters, $h_m(\x)$. They are designed to model the system under study. In the following we propose as measure of similarity the inner product $\mathbf{d}_\x\her\mathbf{d}_\x$ of the difference between complex-valued inputs $\mathbf{d}_\x=\x-\x'$, and use parameterized exponential filters to yield an isotropic and time invariant covariance function. Some examples are provided in the following subsection. This procedure can be applied with other types of filters to model different properties{: periodicity}, other measures of similarity between the inputs, etc. 

\subsubsection{General case}
As \notaRBT{the} first example, we propose to generate the stationary process using filters {$h_{1}(\x)=h_{3}(\x)=v_r\exp{(-{\x\her\x}/{\gamma_r})}$ and $h_{2}(\x)=h_{4}(\x)=v_j\exp{(-{\x\her\x}/{\gamma_j})}$, where $v_r$, $\gamma_r$, $v_j$ and $\gamma_j$ are filter parameters, and the inputs are $\x\in\CC^\d$}. The covariance and pseudo-covariance of the process yield the following functions:{
\begin{align} \LABEQ{K1}
\k(\x,\x\new)&=2v_{r}^{2}\left(\frac{\pi\gamma_r}{2}\right)^{\d}\exp\left(-\frac{\mathbf{d}_\x\her\mathbf{d}_\x}{2\gamma_r}\right)\nonumber\\&+2v_{j}^{2}\left(\frac{\pi\gamma_j}{2}\right)^{\d}\exp\left(-\frac{\mathbf{d}_\x\her\mathbf{d}_\x}{2\gamma_j}\right),
\\
\psc\k(\x,\x\new)&=2v_{r}^{2}\left(\frac{\pi\gamma_r}{2}\right)^{\d}\exp\left(-\frac{\mathbf{d}_\x\her\mathbf{d}_\x}{2\gamma_r}\right)\nonumber\\&-2v_{j}^{2}\left(\frac{\pi\gamma_j}{2}\right)^{\d}\exp\left(-\frac{\mathbf{d}_\x\her\mathbf{d}_\x}{2\gamma_j}\right)\nonumber\\
&+4\j v_{r}v_{j}\left(\frac{\pi\gamma_r\gamma_j}{\gamma_r+\gamma_j}\right)^{\d}\exp\left(-\frac{\mathbf{d}_\x\her\mathbf{d}_\x}{\gamma_r+\gamma_j}\right).\LABEQ{pK1}
\end{align}}
The three real functions {$\k\rrrr(\x,\x\new)$, $\k\jjjj(\x,\x\new)$} and {$\k\rrjj(\x\new,\x)$} of the complex inputs {$\x\in\CC^\d$} are easily identified in this example. 
Note that the covariance \EQ{K1} is real-valued while the pseudo-covariance \EQ{pK1} is complex-valued. This is due to the fact that $\K\jjrr=\K\rrjj\trs=\K\rrjj$ for the second order stationary process generated with the filters. 

\subsubsection{Independent real and imaginary parts}
{In this scenario the cross-covariance between real and imaginary parts is null, $\K\rrjj=\matr{0}$, and we should set $k\rrjj(\x\new,\x)=0$. We can use the same filters proposed for the general case but with the following change of sign: $h_{4}(\x)=-h_{2}(\x)$. Therefore, the covariance function remains as in \EQ{K1}, while the pseudo-covariance function is as in \EQ{pK1} but with the imaginary part equal to zero.}

\subsubsection{Proper case with $k\rrjj(\x\new,\x)=0$}
 In the proper case {$\psc\k(\x,\x\new)=0$ and {if in addition $k\rrjj(\x\new,\x)=0$}},  we use the function in \EQ{K1} {that} yields a simple real covariance function:
\begin{align} \LABEQ{K2}
\k({\x,\x}\new)&=v\exp\left(-{{\mathbf{d}\her_\x\mathbf{d}_\x}/\gamma}\right).
\end{align}

\subsubsection{Proper case with $k\rrjj(\x\new,\x)\neq0$}
This scenario arises when $\K\jjrr$ is skew-symmetric. The kernel functions in \EQ{RK} and \EQ{indepCGK} could be used in this case. However, the first one involves some quite particular similarity properties with exponential {growth} for some pair of points. The second {assumes} constant similarity for distant points as long as they have same real and imaginary parts. We develop a complex-valued function $\k$ to model a correlation between the real part of the process and a displaced or translated imaginary part, with displacement given by $\boldsymbol{\mu} \in \CN^\d, \boldsymbol{\mu}\neq\vect{0}$. $\K\jjrr$ is skew-symmetric if there is also a correlation between the real part and a displaced imaginary part when the displacement is given by $-\boldsymbol{\mu}$, and this correlation has the same value with opposite sign. To model this {behavior,} 
we propose now filters $h_{1}({\x})=h_{3}({\x})=v_r\exp{(-{{\mathbf{d}\her_\x\mathbf{d}_\x}}/{\gamma})}$, $h_{2}({\x})=v_j\exp{(-{{(\x- \boldsymbol{\mu})\her(\x- \boldsymbol{\mu})}}/{\gamma})}$, and $h_{4}({\x})=-v_j\exp{(-{{(\x+\boldsymbol{\mu})\her(\x+\boldsymbol{\mu})}}/{\gamma})}$. 
The covariance function yields
\begin{align}\LABEQ{kerfil}
{\k(\x, \x\new)}&={v_A}\exp\left(-\frac{{\mathbf{d}\her_\x\mathbf{d}_\x}}{2\gamma}\right)\nonumber\\
&+{\j v_B}\left(\exp\left(-\frac{{(\mathbf{d}_\x-\boldsymbol{\mu})\her(\mathbf{d}_\x-\boldsymbol{\mu})}}{2\gamma}\right)\right.\nonumber\\
&-\left.\exp\left(-\frac{{(\mathbf{d}_\x+\boldsymbol{\mu})\her(\mathbf{d}_\x+\boldsymbol{\mu})}}{2\gamma}\right)\right),
\end{align} 
where ${v_A=2\left(v\rr^{2}+v\jj^{2}\right)\left(\frac{\pi\gamma}{2}\right)^{\d}}$ and ${v_B=2v\rr v\jj\left(\frac{\pi\gamma}{2}\right)^{\d}}$.
{The} pseudo-covariance yields
\begin{align}\LABEQ{pkerfil}
{\psc\k(x, x\new)}&={2}\left(v\rr^{2}-v\jj^{2}\right){\left(\frac{\pi\gamma}{2}\right)^{\d}}\exp\left(-\frac{{\mathbf{d}\her_\x\mathbf{d}_\x}}{2\gamma}\right).
\end{align} 
By setting $v\rr=v\jj$ we get the proper {case}.

\section{Hyperparameters estimation}\LABSEC{ML}
A major advantage of GPR is that the hyperparameters can also be estimated by maximizing the marginal likelihood \cite{Rasmussen06}. From the marginal likelihood \EQ{marlikelihood} we can compute the log marginal likelihood
\begin{align} \label{eq:CMML}
L(\vect{\theta})=\log p\left(\yv|\X\right)=-\frac{1}{2}\aug{\yv}\her\aug{\C}\inv\aug{\yv}-\frac{1}{2}\log\det\aug{\C}-\n\log \pi.
\end{align}
The augmented covariance matrix $\aug{\C}=\aug{\C}(\boldsymbol{\theta})$ can be parameterized in terms of the hyperparameters $\vect{\theta}_{i}$, which are the parameters of the covariance and pseudo-covariance functions and the noise variance and pseudo-covariance. $L\left(\boldsymbol{\theta}\right)$ is a real function of a complex-valued Hermitian matrix. Therefore, in the maximization of \EQ{CMML}, we must seek generalized complex-valued matrix derivatives \cite{Hjorungnes08,Hjorungnes07}. The result for the gradient, as developed in \APEN{Ap2}, is as follows,
\begin{align} \label{eq:gradientefinal}
\frac{\partial {L}}{\partial \theta_i} ={\mathrm{Tr}\left(\left(\aug{\C}^{-1}\aug{\yv}\,\aug{\yv}\her\aug{\C}^{-1}-\aug{\C}^{-1}\right){\frac{\partial {\aug{\C}}}{\partial \theta_i}}\right)}.
\end{align}

In the proper case, when $\psc{\C}=\matr{0}$, the gradient simplifies to
\begin{align} \label{eq:gradientefinalp}
\frac{\partial {L}}{\partial \theta_i} =\notaRBT{2}{\mathrm{Tr}\left(\left(\C^{-1}\yv\subn\yv\subn\her\C^{-1}-\C^{-1}\right){\frac{\partial {\C}}{\partial \theta_i}}\right)}.
\end{align}
%

\section{Experiments}\LABSEC{Exp}
We include three experiments. {First,} {we} evaluate the full CGPR solution against the simpler proper CGPR. Then we illustrate the performance of the {proper CGPR} in an scenario where the covariance is complex-valued. Finally, we face the equalization of \notaRBT{nonlinear} channels to compare to previous solutions. In this last experiment, we use the recursive version of the proper CGPR.

\subsection{\notaRBT{Full CGPR}}

We generate a sample function of a non-proper complex-valued Gaussian process as described in \SSEC{convmeth}-{1)}. {The inputs in this experiment are complex-valued \notaRBT{scalars}, i.e., $\x=x\in \CN$, {to easily represent} the sample function of the process in a figure}. {We have set} $h_{1}(x)=h_{3}(x)=0.1\exp{(-{x^*x}/{0.6})}$ and $h_{2}(x)=h_{4}(x)=0.05\exp{(-{x^*x}/{1.5})}$. 
The covariance function {is given by} \EQ{K1}, $\k(x,x\new)=0.006\pi\exp\left(-({d}_\x^*{d}_\x)/{1.2}\right)+0.0038\pi\exp\left(-({{d}_\x^{*}{d}_\x})/{3}\right)$,
while the pseudo-covariance function is as in \EQ{pK1}, 
$\psc\k(x,x')=0.006\pi\exp\left(-({d}_\x^*{d}_\x)/{1.2}\right)-0.0038\pi\exp\left(-({{d}_\x^{*}{d}_\x})/{3}\right)+\j{0.0086\pi}\exp\left(-({d}_\x^*{d}_\x)/2.1\right)$.
As an example, the real part of a sample function obtained, {$f(x)$}, is shown in \FIG{fig13_2}. {The imaginary part is shown in \FIG{fig13_22}}. Gaussian noise with variance $\sigma_{\epsilon}^2$ and {pseudo-variance} $\rho\sigma_{\epsilon}^2$ is added to represent measurement uncertainty, where 
$\rho=0.8\exp(\j 3\pi/2)$ and $\sigma^2_{\epsilon}$ is set to be 25 dB below the variance of the sample function {(signal-to-noise \notaRBT{ratio}, SNR = 25 dB)}.
Then, we randomly choose $\n=500$ noisy training samples and learn the sample function of the process by using the predictive mean \EQ{meanfnew}, variance \EQ{varfnew} and pseudo-variance \EQ{pvarfnew}. The training samples used are marked as circles in {Figs. \ref{fig:fig13_2} and \ref{fig:fig13_22}. The real part of the predictive mean is shown in \FIG{fig13_2} (bottom) while the imaginary part is shown in \FIG{fig13_22} (bottom).} The mean squared error (MSE) of the estimation is $10\log_{10}$(MSE)$ =-8.2$ dB, computed for $10^4$ inputs in this example.


%

%
\begin{figure}[tb!]
\begin{center}
\includegraphics[width=8cm, draft=false]{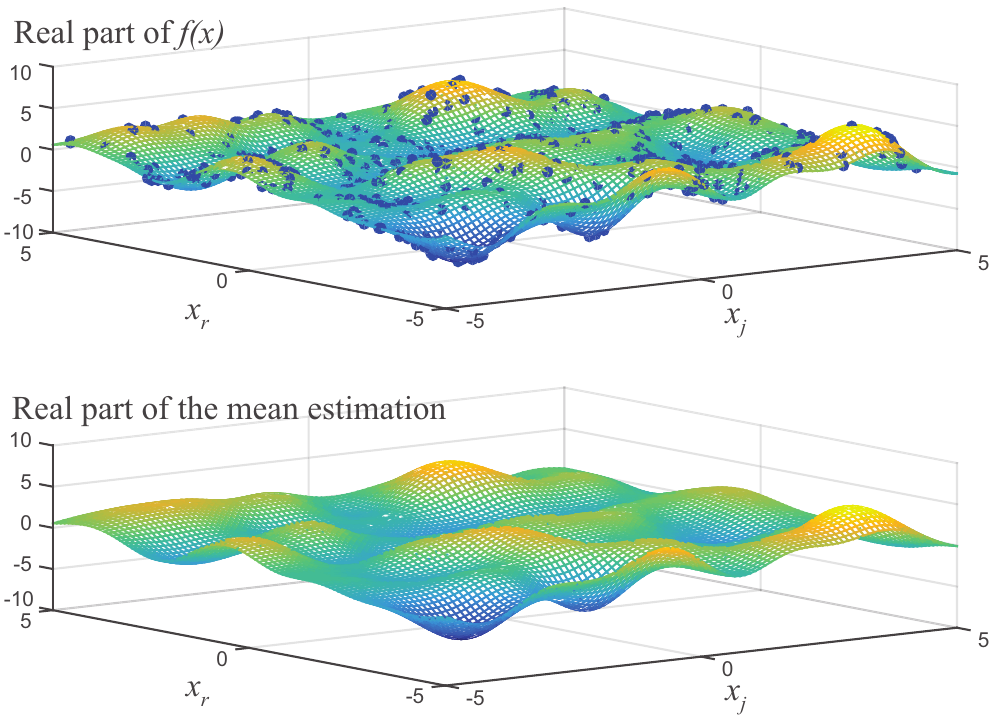}
\end{center}
\caption{{Real part of the sample function $f(x)$ of the process (top) and real part of the mean estimation (\ref{eq:meanfnew}) (bottom) versus the real and imaginary parts of the input, $x_{r}$ and $x_{j}$. The training samples are depicted as blue circles.}}
\LABFIG{fig13_2}
\end{figure}
\begin{figure}[tb!]
\begin{center}
\includegraphics[width=8cm, draft=false]{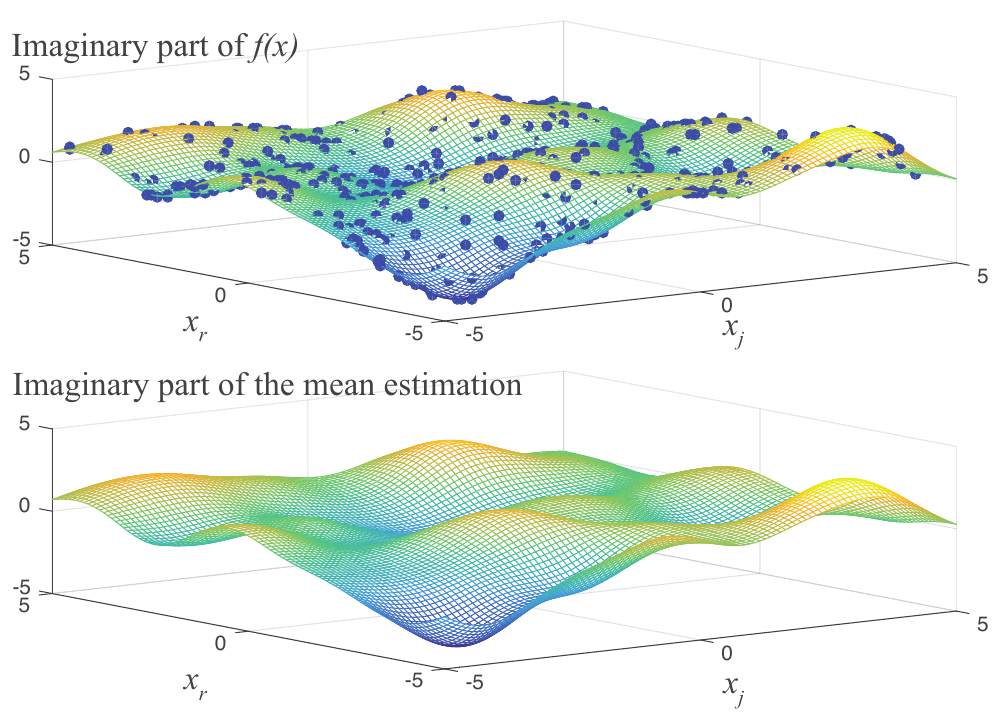}
\end{center}
\caption{{Imaginary part of the sample function $f(x)$ of the process (top) and real part of the mean estimation (\ref{eq:meanfnew}) (bottom) versus the real and imaginary parts of the input, $x_{r}$ and $x_{j}$. The training samples are depicted as blue circles.}}
\LABFIG{fig13_22}
\end{figure}

\begin{figure}[tb!]
\begin{center}
\includegraphics[width=8.6cm, draft=false]{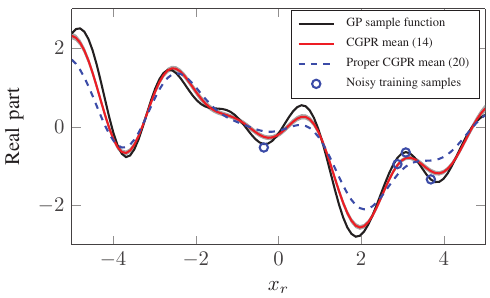}
\end{center}
\caption{Real parts of the sample function of the process $f(\x)$, the predictive CGPR mean (\ref{eq:meanfnew}), and the predictive mean for the proper CGPR case \EQ{mediaCPGP}, versus the real part of the input $x_{r}$, for $x_{j}=-0.1515$.}
\LABFIG{fig14}
\end{figure}

\begin{figure}[tb!]
\begin{center}
\includegraphics[width=8.6cm, draft=false]{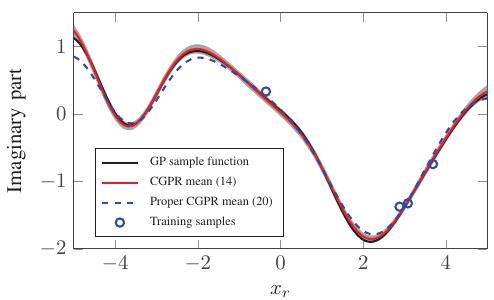}
\end{center}
\caption{Imaginary parts of the sample function of the process $f(\x)$, the predictive CGPR mean (\ref{eq:meanfnew}), and the predictive mean for the proper CGPR case \EQ{mediaCPGP}, versus the real part of the input $x_{r}$, for $x_{j}=-0.1515$.}
\LABFIG{fig15}
\end{figure}

\begin{figure}[htb!]
\begin{center}
\includegraphics[width=8.6cm, draft=false]{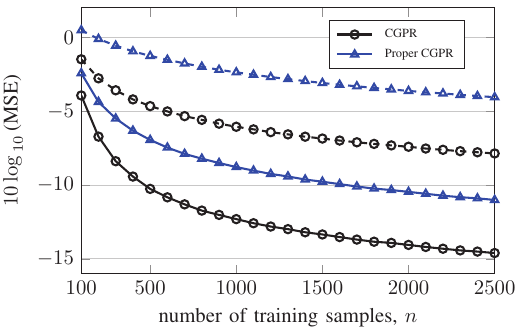}
\end{center}
\caption{Averaged $10\log_{10}${(MSE)} versus the number of training samples for the predictive CGPR mean (\ref{eq:meanfnew}) and the proper CGPR case (\ref{eq:mediaCPGP}). Solid line: SNR = 25 dB. Dashed line: SNR = 10 dB.}
\LABFIG{widelyvsproper}
\end{figure}

The predictive capability of the complex GPR in (\ref{eq:meanfnew}) is compared with that of the proper case in \EQ{mediaCPGP}. The mean squared error (MSE) of the estimation for the proper case is $10\log_{10}$(MSE)$ =-4.67$ dB in this example. We show in Figs. \ref{fig:fig14} and \ref{fig:fig15} randomly chosen slices of the sample function in Figs. \ref{fig:fig13_2} and \ref{fig:fig13_22}. The imaginary part of the input was fixed to $x_{j}=-0.1515$. In \FIG{fig14} we include the real part of the prediction in (\ref{eq:meanfnew}) and the grey shaded area that represents the point-wise mean plus and minus two times the standard deviation. The mean of the prediction for the proper case in (\ref{eq:mediaCPGP}) is plotted as a dashed line. In \FIG{fig15} we include the imaginary part. The general CGPR prediction is always closer to the actual value of  {$f(x)$} than the prediction for the proper case, as expected, since the general CGPR also uses the information of the pseudo-covariance. This prediction improvement is highlighted in \FIG{widelyvsproper}, where we compare the mean squared error for both estimations along the number of training samples. A higher noise case (SNR = 10 dB) is also included. Results are the average of 100 simulated trials for each case in this example. 
The proposed complex GPR performs better {than the proper CGPR}, with a remarkable reduction in the number of  training samples. {We achieve the same MSE of {$-10$ dB} with a sizable reduction in the number of training examples, i.e. from $1500$ to $500$ for an SNR of $25$ dB.} 


\subsection{\notaRBT{Proper CGPR}}
To illustrate the performance of the hyperparameter estimation we face the learning of a proper complex Gaussian process with complex-valued covariance function. In this scenario $\K\jjrr$ is skew-symmetric. {Again, the inputs in this experiment are complex-valued scalar, i.e., $\x=x\in \CN$, for representation purposes}. We use the covariance function in \EQ{kerfil}, 
with {$v_A=2$, $v_B=1$,} 
$\gamma=1.125$ and $\mu=2+2\j$, to generate a sample function of the process, $f(x)$. 
{The real part of the sample function is shown in \FIG{fig2re}, while the imaginary part is {depicted} in  \FIG{fig2}}. Circular complex Gaussian noise with $\sigma_{\epsilon}=0.1$ is added to represent measurement uncertainty and $\n=200$ training noisy samples are randomly chosen as training data. 
The maximization of the log marginal likelihood in (\ref{eq:CMML}) using (\ref{eq:gradientefinalp}) yields the following estimated values of the hyperparameters: {$\hat{v}_A=2.1169$, $\hat{v}_B=1.1425$, $\hat\gamma=1.1373$, $\hat\mu=1.9371 +  \j1.9983$, and $\hat\sigma_{\epsilon}=0.0968$}. Then, the mean (\ref{eq:mediaCPGP}) and variance (\ref{eq:varCPGP}) of the predictive distribution are found using the training samples and the estimated values of the hyperparameters. {The real part of the predictive mean (\ref{eq:mediaCPGP}) is depicted in \FIG{fig2re} (bottom), while the imaginary part is depicted in \FIG{fig2} (bottom)}. The MSE of the estimation is {$-13.8807$} dB. We include in {Figs. \ref{fig:fig3} and \ref{fig:fig4} randomly chosen slices of the sample function. Fig. \ref{fig:fig3} shows the real part of the sample function and the real part of the prediction (\ref{eq:mediaCPGP}) versus the real part of the input, for $x\jj=3.4684$. Fig. \ref{fig:fig4} shows the imaginary part of the sample function, and the imaginary part of the prediction (\ref{eq:mediaCPGP}) versus the real part of the input, for $x\jj=-5.4430$. The training samples are depicted as blue circles. Also, four instances of the posterior are plotted in both Figs. \ref{fig:fig3} and \ref{fig:fig4}}.

{To complete the analysis we show in Figs. \ref{fig:fig47b} and \ref{fig:fig47a} the MSE of the estimation for each hyperparameter by maximizing the log marginal likelihood in (\ref{eq:CMML}) using (\ref{eq:gradientefinalp}) under different settings. Fig. \ref{fig:fig47b} shows the MSE versus the SNR for a fixed number of training samples $\n=200$, while Fig. \ref{fig:fig47a} shows the MSE versus the number of training samples for a fixed SNR of $16$ dB. The MSE is the averaged value for $100$ independent trials. Finally, we compare in Fig. \ref{fig:fig46} the MSE of the learning when these estimated hyperparameters are used to calculate the prediction (\ref{eq:mediaCPGP}) with the prediction calculated with the true hyperparameters. In Fig. \ref{fig:fig46} (top) the number of training samples was fixed to $\n=200$, while in Fig. \ref{fig:fig46} (bottom) the SNR was fixed to $16$ dB. The MSE curves are very close in both cases. }
%
%
\begin{figure}[tb!]
\begin{center}
\includegraphics[width=8.6cm, height=5cm,draft=false]{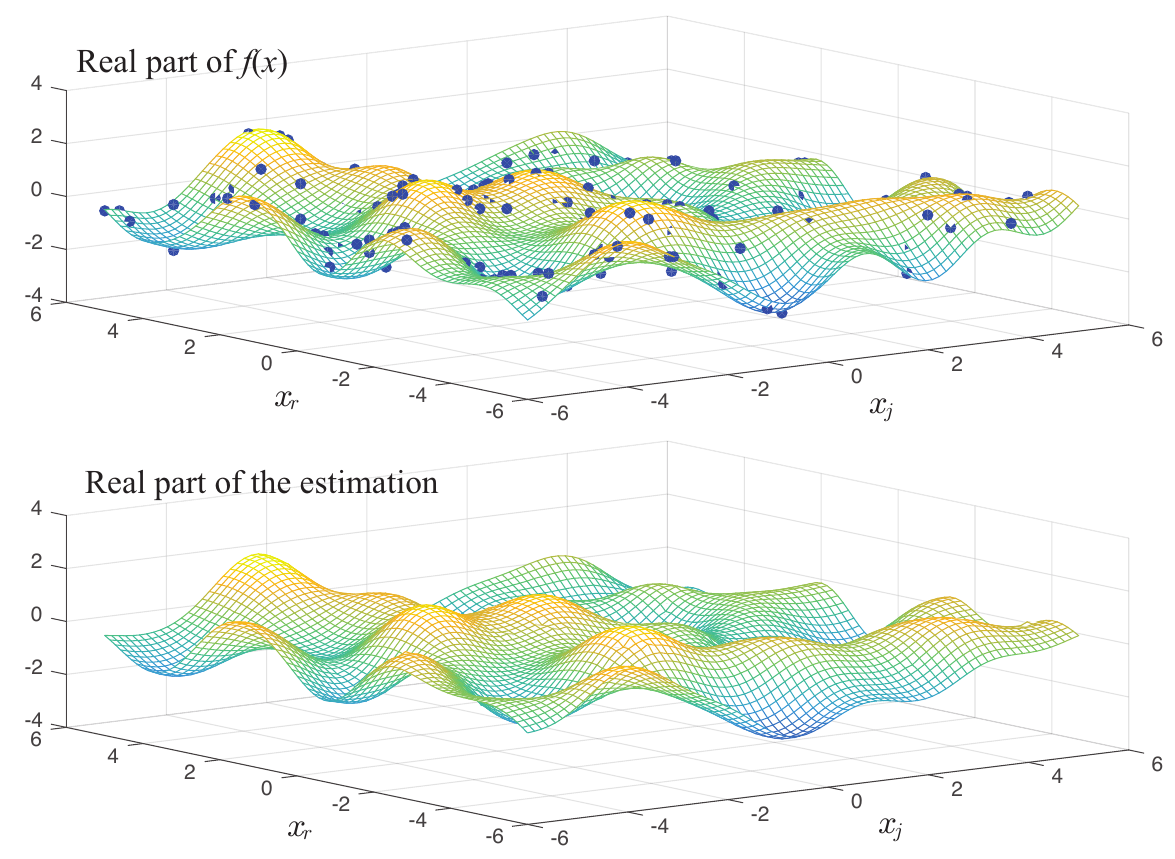}
\end{center}
\vspace*{-.6cm}
\caption{{Real part of the sample function $f(x)$ of the process (top) and real part of the mean estimation (\ref{eq:mediaCPGP}) (bottom) versus the real and imaginary parts of the input. The training samples are depicted as blue circles.}}
\LABFIG{fig2re}
\end{figure}
\begin{figure}[tb!]
\begin{center}
\includegraphics[width=8.6cm, height=5cm,draft=false]{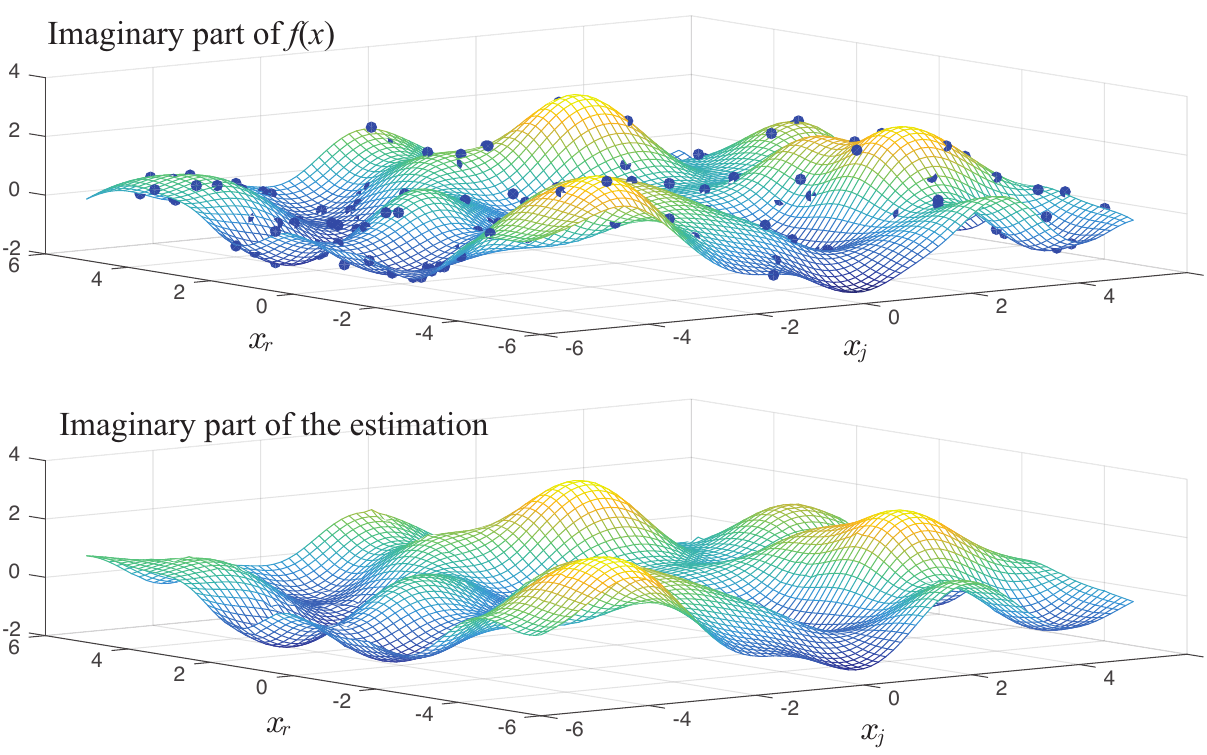}
\end{center}
\vspace*{-.6cm}
\caption{Imaginary part {of} the sample function {$f(x)$} of the process (top) and imaginary part of the mean estimation (\ref{eq:mediaCPGP}) (bottom) versus the real and imaginary parts of the input. The training samples are depicted as blue circles.}
\LABFIG{fig2}
\end{figure}
\begin{figure}[tb!]
\begin{center}
\includegraphics[width=8.5cm, height=5cm,draft=false]{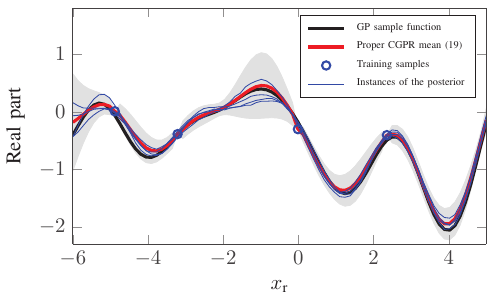}
\end{center}
\vspace*{-.6cm}
\caption{Real parts of the output and the predictive mean (\ref{eq:mediaCPGP}) versus the real part of the input $x\rr$ for {$x\jj=3.4684$}. Training samples are depicted as blue circles. Four instances of the posterior  are also plotted.}
\LABFIG{fig3}
\end{figure}
\begin{figure}[htb!]
\begin{center}
\includegraphics[width=8.6cm, height=5cm,draft=false]{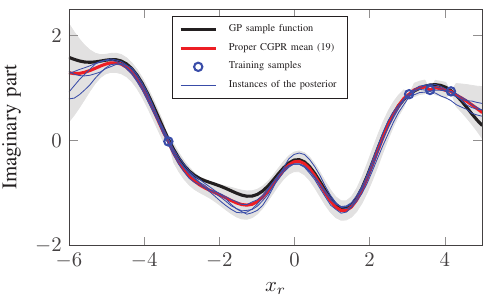}
\end{center}
\vspace*{-.6cm}
\caption{Imaginary parts of the output and the predictive mean (\ref{eq:mediaCPGP}) versus the real part of the input $x\rr$ for {$x\jj=-5.4430$}. Training samples are depicted as blue circles. Four instances of the posterior are also plotted.}
\LABFIG{fig4}
\end{figure}

\begin{figure}[htb!]
\begin{center}
\includegraphics[width=8.6cm, draft=false]{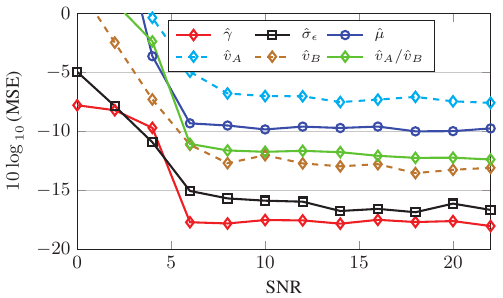}
\end{center}
\caption{{Hyperparameters learning curve versus the SNR using 200 training samples.}}
\LABFIG{fig47b}
\end{figure}
\begin{figure}[htb!]
\begin{center}
\includegraphics[width=8.6cm, draft=false]{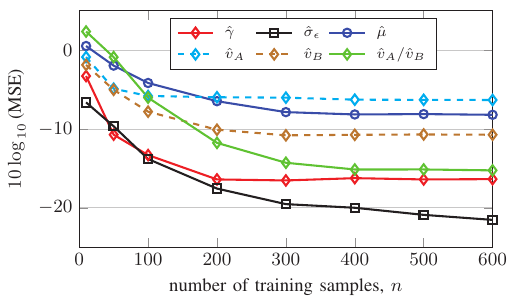}
\end{center}
\caption{{Hyperparameters learning curve versus the number of training samples for $SNR=16$.}}
\LABFIG{fig47a}
\end{figure}

\begin{figure}[htb!]
\begin{center}
\includegraphics[width=8.5cm, height=5cm,draft=false]{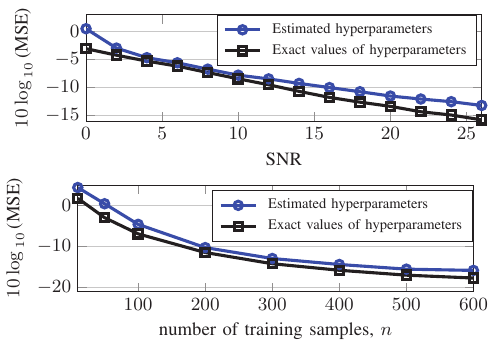}
\end{center}
\caption{{Predictive MSE versus the SNR for 200 training samples (above), and Predictive MSE versus the number of training samples for $SNR=16$ (below).}}
\LABFIG{fig46}
\end{figure}

{We include in Figs. \ref{fig:fig47b} and \ref{fig:fig47a} the MSE of the estimation of the ratio $v_A/v_B$. This ratio is more important than the actual value of $v_A$ or $v_B$ for the estimation of the mean in \EQ{mediaCPGP}. 
{The} ratio $v_A/v_B$ and $\sigma_\epsilon$ are responsible for the amplitude accuracy of the estimation. As shown in Figs. \ref{fig:fig47b} and \ref{fig:fig47a} the MSE for $\hat{v}_A/\hat{v}_B$ and $\hat\sigma_\epsilon$ are low and, therefore, the MSE  of the function estimation is {low}, as shown in Fig. \ref{fig:fig46}.}


\subsection{Nonlinear channel equalization}
The performance of the proposed complex GPR is tested in the context of the nonlinear channel equalization task in \cite{Bouboulis11} and \cite{Bouboulis12}. Two nonlinear channels are considered. Both \notaRBT{channel models} consist of a linear filter $t(n)=(-0.9+0.8\j)\cdot s(n)+(0.6-0.7\j)\cdot s(n-1)$ and {a nonlinear function. The linear filter represents a communication channel with memory, while the nonlinear function represents the effect of nonlinear circuits, such as amplifiers}. The nonlinearity is $q(n)=t(n)+(0.1+0.15\j)\cdot t^{2}(n)+ (0.06+0.05\j)\cdot t^{3}(n)$ for the first case (labeled as {\it soft nonlinear channel}), and $q(n)=t(n)+(0.2+0.25\j)\cdot t^{2}(n)+ (0.12+0.09\j)\cdot t^{3}(n)$ for the second case (labeled as {\it strong nonlinear channel}). The input signals are $s(n)=0.70(\sqrt{1-\rho^{2}}X(n)+\j\rho Y(n))$,  and $X(n)$ and $Y(n)$ were Gaussian random variables. The input signals are circular for $\rho=1/\sqrt{2}$ and highly noncircular if $\rho$ approaches $0$ or $1$. At the receiver end of the channel, the signal $q(n)$ was corrupted by additive white circular Gaussian noise {with a SNR of} 16 dB, as in \cite{Bouboulis11}.

The aim of the channel equalization task is to construct an inverse filter, which acts on the received signal $r(t)$ and reproduces the original input signal $s(n)$ as close as possible. To this end, the inputs to the equalizer are the sets of samples $\vect{x}(n)=[r(n+D),r(n+D-1),\cdots,r(n+D-L+1)]^{\top}$, where $L>0$ is the filter length and $D$ is the equalization time delay. Experiments are conducted as in \cite{Bouboulis11} and \cite{Bouboulis12}, where $L=5$ and $D=2$, on a set of 5000 samples of the input signal considering both the circular and the noncircular ($\rho=0.1$) cases and the ({\it soft} and {\it strong}) nonlinear channels. In all cases the results are averaged over 500 trials where the input signal samples $s(n)$ and noise are randomly generated. 

The performance of our proposal is compared with the NCKLMS2 algorithm in \cite{Bouboulis11}, the ACKLMS algorithm in \cite{Bouboulis12} and the iCKLMS in \cite{Tobar12}. Both the NCKLMS2 and ACKLMS algorithms use the complex Gaussian kernel in \EQ{RK}. The iCKLMS is as the NCKLMS2 algorithm but using the independent kernel (\ref{eq:indepCGK}) with $\kappa_{\Rext}$ being the real Gaussian kernel. We use the code available in \cite{boubouliscode} to {run these} algorithms.  All the parameters required for the NCKLMS2 and the ACKLMS algorithms ($\gamma$ in kernel and step update parameter) are set to the values described in \cite{Bouboulis11} and \cite{Bouboulis12}, except for the {\it strong nonlinear channel} noncircular case, where in order to ensure convergence we increase $\gamma$ to {$\gamma=400$} for both algorithms. For the iCKLMS,  {$\gamma=25$} and the step update parameter is 1/8 (except for the {\it strong nonlinear channel} noncircular case, where it is reduced to 1/16), tuned for the best possible result.  For the three algorithms the novelty criterion \notaRBT{\cite{Liu10,Platt91}} is used for the sparsification with $\delta_{1}=0.15$ and $\delta_{2}=0.2$, as in \cite{boubouliscode}.

\begin{figure}[tb!]
\begin{center}
\includegraphics[width=8.6cm, draft=false]{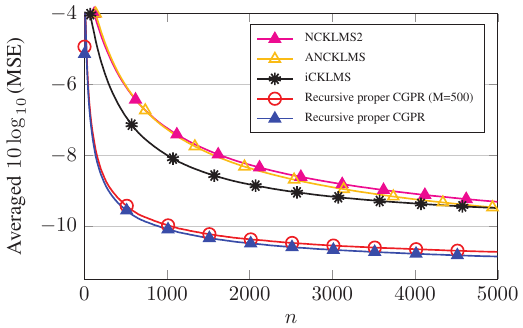}
\end{center}
\caption{Averaged MSE along $\n$ for NCKLMS2, ANCKLMS, iCKLMS, the recursive proper CGPR and the recursive proper CGPR with M=500 basis for the soft nonlinear channel equalization problem and the circular input case.}
\LABFIG{fig5}
\end{figure}
\begin{figure}[tb!]
\begin{center}
\includegraphics[width=8.6cm, draft=false]{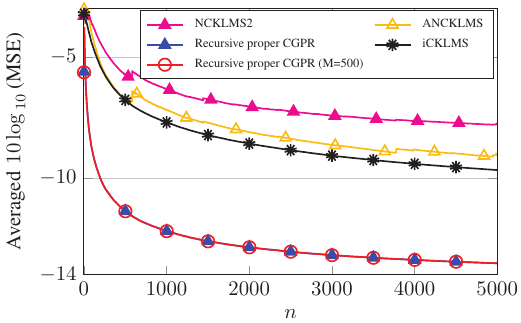}
\end{center}
\caption{Averaged MSE along $\n$ for NCKLMS2, ANCKLMS, iCKLMS, the recursive proper CGPR and the recursive proper CGPR with M=500 basis for the strong nonlinear channel equalization and the noncircular input case ($\rho=0.1$).}
\LABFIG{fig8}
\end{figure}
\begin{figure}[tb!]
\begin{center}
\includegraphics[width=8.6cm, draft=false]{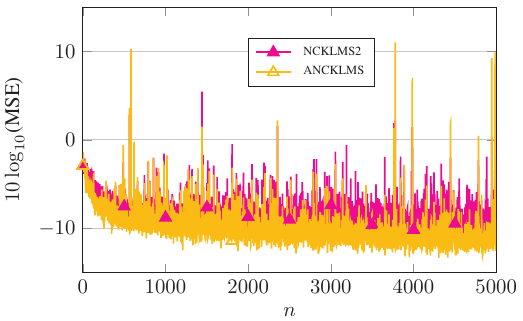}
\end{center}
\caption{MSE along $\n$ for NCKLMS2 and ANCKLMS for the strong nonlinear channel equalization problem for the noncircular input case ($\rho=0.1$). }
\LABFIG{fig82}
\end{figure}
\begin{figure}[tb!]
\begin{center}
\includegraphics[width=8.6cm, draft=false]{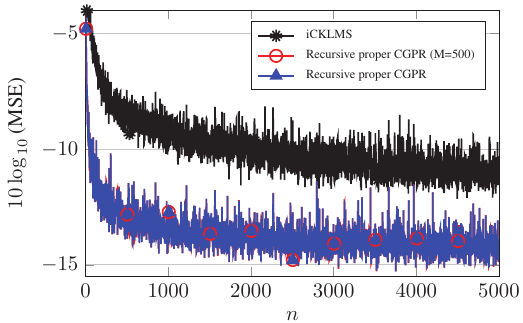}
\end{center}
\caption{MSE along $\n$ for iCKLMS, the recursive proper CGPR and the recursive proper CGPR with M=500 basis for the strong nonlinear channel equalization problem for the noncircular input case ($\rho=0.1$). }
\LABFIG{fig83}
\end{figure}

We design a CGPR solution as follows. \notaRBT{The CGPR outputs here are the signals $s(n)$. 
Note first that the real and the imaginary parts of $s(n)$ are generated independently and therefore have null cross-covariances, $\K\rrjj=\matr{0}$. In such a case, $\K=\K\rrrr+\K\jjjj$ in \EQ{covK} and $\psc\K=\K\rrrr- \K\jjjj$ in \EQ{pcovK}, and it is not necessary to use complex-valued covariance functions, as it was explained in  \SEC{Cov}. Both $\K\rrrr$ and $\K\jjjj$ can be obtained from a real kernel as in \EQ{K2}. Also, in this equalization application, when trying to set the values of the hyperparameters, we found that independently of the factor $\rho$ the best solution is achieved with $\K\rrrr=\K\jjjj$. Therefore, in this scenario the general case, the full CGPR, reduces to the proper CGPR, as we can set $\psc\K=\K\rrrr- \K\jjjj=\vect{0}$.}
%
Hence, the proper version of the CGPR in \EQ{mediaCPGP} suffices and we propose a real-valued covariance function as the one in \EQ{K2}. However, since both the NCKLMS2 and the ACKLMS are online sequential algorithms we have to use an online algorithm for the proper CGPR also, in order to provide a fair comparison. In \cite{Vaerenbergh12}, {the authors provide} a Bayesian derivation for the kernel recursive least-squares algorithm and a criterion to remove the least relevant basis (the set of inputs at which the joint posterior is available, i.e., the training samples in our setting). Since the {pseudo}-covariance cancels, the method in \cite{Vaerenbergh12} can be easily adapted to the proper CGPR case in order to yield a recursive proper CGPR. The objective is to infer the conditional distribution $p\left(\fv_{\n+1}| \tset, \x\new,y\new\right)$ of $\fv_{\n+1}=[\fv_{\n}^{\top},f\new]^{\top}$ given \notaRBT{the training set $\tset\subn=\{[\x_1, . . . , \x_\n],[{\y_1}, . . . , {\y_\n}]^{\top}\}$ and a new input $\x\new=\x(\n+1)$ with corresponding output $y\new=y(\n+1)$}, where $f\new=f(\x\new)$. 
We apply the recursive proper CGPR with basis removal criterion using $M=500$ bases where the first $250$ samples were used for the hyperparameters estimation by maximizing the log marginal likelihood in (\ref{eq:CMML}) using (\ref{eq:gradientefinalp}). As reference we also include the recursive proper CGPR solution without basis removal criterion with $1000$ randomly chosen samples among the total of $5000$ used to find a better estimation of the hyperparameters. 
Note that the number of bases used by the NCKLMS2, ACKLMS or iCKLMS algorithms with the novelty sparsification criterion grew above $2000$ in these experiments, and therefore the choice of $M=500$ bases is far below that number. 

We show in Figs. \ref{fig:fig5} and \ref{fig:fig8} the averaged MSE along the input samples for the NCKLMS2, the ACKLMS, the iCKLMS and the two recursive proper CGPR algorithms (with and without basis removal criterion), for the {\it soft} nonlinear channel circular and the {\it strong} nonlinear with $\rho=0.1$ cases. The MSE value depicted for each sample is the averaged MSE for all previous outputs, as in \cite{Bouboulis11,boubouliscode}.  
It can be observed in the figures the remarkable good results of the recursive proper CGPR in all cases, even with only $250$ bases used for the hyperparameters estimation and $M=500$ bases used for the prediction, with the additional advantage of the estimation of the hyperparameters from the samples, avoiding cross-validation. This solution is very close to the proper CGPR approach used as reference.
\notaRBT{The raw MSE, i.e., not averaged, for the {\it soft} nonlinear channel circular and the {\it strong} nonlinear with $\rho=0.1$ cases are shown in Figs. \ref{fig:fig82} to \ref{fig:fig85}. }

By using the proposed solution we avoid convergence problems found in the learning process of both the NCKLMS2 and ACKLMS algorithms. These problems 
can be observed if the MSE is not averaged for the previous outputs. As examples, we provide in Fig. 
\ref{fig:fig82} the same results that were included in Fig. 
\ref{fig:fig8} for the NCKLMS2 and ACKLMS algorithms, but now the MSE is not averaged.  Notice that those algorithms are not able to provide a good prediction for some outputs, with MSE peak values above $5$ dB. This is not the case for the iCKLMS and the recursive proper CGPR, as can be observed in Fig.
\ref{fig:fig83}. We believe that the NCKLMS2 or ACKLMS algorithms fail to provide a good prediction for some outputs because of the kernel they use. The independent kernel in the iCKLMS algorithm seems a better choice than the complex Gaussian kernel. However, the proper CGPR with the real kernel provides, by far, the best solution.

\begin{figure}[tb!]
\begin{center}
\includegraphics[width=8.6cm, draft=false]{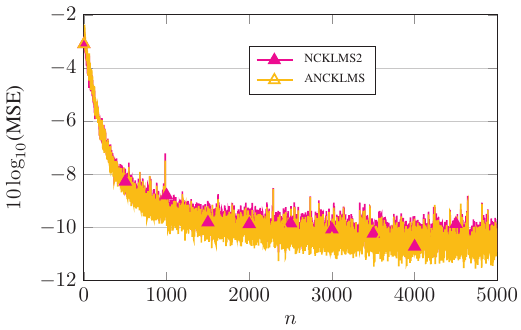}
\end{center}
\caption{\notaRBT{MSE along $\n$ for NCKLMS2 and ANCKLMS for the soft nonlinear channel equalization problem for the circular input case.} }
\LABFIG{fig84}
\end{figure}
\begin{figure}[tb!]
\begin{center}
\includegraphics[width=8.6cm, draft=false]{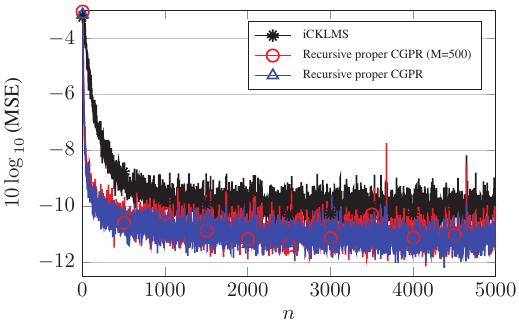}
\end{center}
\caption{\notaRBT{MSE along $\n$ for iCKLMS, the recursive proper CGPR and the recursive proper CGPR with M=500 basis for the soft nonlinear channel equalization problem for the circular input case.} }
\LABFIG{fig85}
\end{figure}



%
\section{Conclusions}

Regression in the complex-valued case has been addressed by dealing with real and imaginary parts independently, using a straightforward extension of the real case or learning a vector with real and imaginary parts stacked. {However, in these approaches the design of the kernels remains an open problem and the complex-valued formulation is lost. On the other hand, the straightforward adaptation of the real case to the complex one corresponds to the proper case, and is not able to deal with any scenario.} In this paper we present a new approach based on the results for complex-valued Gaussian processes. To the best of our knowledge this is the first tool working in the complex field, suitable for any scenario.

We exploit the GPR framework to provide a full statistical description of the general complex-valued solution. {We highlight the importance of the pseudo-covariance term,} and the mean and covariance of the posterior are developed. Only when the pseudo-covariance cancels, the method simplifies to the proper case. We develop the optimization of the marginal likelihood to estimate the hyperparameters, by taking into account generalized complex-valued matrix derivatives. The selection or design of the covariance function or kernel is also an important issue that we deal with in this paper. {We analyze the {terms} in the real and imaginary parts of the covariance and pseudo-covariance functions, and their symmetries.} We review some previous {proposals}, and the way these kernels measure similarity between the complex-valued inputs. We propose a {more general} method to design the covariance and pseudo-covariance functions from filters. 
We highlight the importance of focusing on the properties of the covariance and pseudo-covariance for the problem at hand to get the simplest solution needed. In particular, {when the function we would like to fit does not have null pseudo-covariance, then the general CGPR formulation provides better results. On the other hand, if the pseudo-covariance is null, the simpler proper CGPR is enough. Two experiments are included to illustrate these facts, showing the learning of non-proper and proper models, along with the learning of the hyperparameters. Also,} when the cross-covariance between the real and imaginary parts is symmetric or null, there is no need for a complex-valued covariance function. These developments are in the line of solving the equalization of nonlinear channels in the experiments section, {where we propose a real-valued covariance function while previous solutions {use} a complex-valued one}. \notaRBT{We apply a recursive version of the proper CGPR with a basis selection criterion and compare the results to previous approaches. The recursive proper CGPR yields a remarkable reduction of the MSE, up to 4 dB, and with a number of bases that is less than $25$\% of the number required for previous approaches. Also, the proper CGPR approach allows us to learn the hyperparameters from the data, so there is no need to set them by extensive search or cross-validation techniques.} 




\appendix
\subsection{Design of a Complex Covariance Function}\LABAPEN{Apen1}
We follow here a procedure similar to that in \cite{Boyle05}. Consider two independent, real, Gaussian white noise processes, with zero mean and unit variance, $S\rr(\x)$ and $S\jj(\x)$, where $\x\in\CN^\d$, producing an output $U(\x)$ defined by the sum of convolutions
\begin{align}
U(\x)&=\left(h_{1}(\x)+\j h_{2}(\x)\right) \star S\rr(\x)+\left(h_{3}(\x)+\j h_{4}(\x)\right) \star S\jj(\x)\nonumber\\
&=\sum_{m=1}^4 \lambda_{m}h_m(\x) \star S_m(\x),
\end{align}
where $\lambda_{1}= \lambda_{3}={1}$ and $\lambda_{2}= \lambda_{4}={\j}$, $S_1(\x)=S_2(\x)= S\rr(\x)$, and $S_3(\x)=S_4(\x)= S\jj(\x)$.

The covariance of $U(\x)$ is derived as follows:
\begin{align} \LABEQ{app2}
\C_{U}(\x,\x\new)&=\E\left[U(\x)U^{*}(\x\new)\right]\nonumber\\
&=\E\left[\sum_{m=1}^4\int_{\CN^\d}{\lambda_{m}h_m(\boldsymbol{\alpha})S_m(\x-\boldsymbol{\alpha})\dfd{\d}{\boldsymbol{\alpha}}}\right.\nonumber\\&\cdot\left.\sum_{n=1}^4\int_{\CN^\d}{\lambda^{*}_{n}h_n^{*}(\boldsymbol{\beta})S_n(\x\new-\boldsymbol{\beta})\dfd{\d}{\boldsymbol{\beta}}}\right]
\nonumber\\
&=\sum_{m=1}^4\sum_{n=1}^4\left\{\int_{\CN^\d}\int_{\CN^\d}\lambda_{m}\lambda^{*}_{n}h_m(\boldsymbol{\alpha})h_n^{*}(\boldsymbol{\beta})\right.\nonumber\\&\cdot\left.\E\left[S_m(\x-\boldsymbol{\alpha})S_n(\x\new-\boldsymbol{\beta})\right]\dfd{\d}{\boldsymbol{\alpha}}\dfd{\d}{\boldsymbol{\beta}}\right\}.
\end{align}
Processes $S_m(\x-\boldsymbol{\alpha})$ and $S_n(\x\new-\boldsymbol{\beta})$ covary only if $m,n\in\{1,2\}$ or $m,n\in\{3,4\}$, and $(\x-\boldsymbol{\alpha})=(\x\new-\boldsymbol{\beta})$. In such cases, $\E\left[S_m(\x-\boldsymbol{\alpha})S_n(\x\new-\boldsymbol{\beta})\right]=\delta(\boldsymbol{\alpha}-(\x-\x\new+\boldsymbol{\beta}))=\delta(\boldsymbol{\alpha}-(\mathbf{d}_\x+\boldsymbol{\beta}))$, \notaRBT{where $\delta(\cdot)$ is the Dirac delta function,} and the integrals in \EQ{app2} yield
\begin{align}\LABEQ{efes}
&f_{mn}(\mathbf{d}_\x)=\nonumber\\
&\int_{\CN^\d}\int_{\CN^\d}\lambda_{m}\lambda^{*}_{n}h_m(\boldsymbol{\alpha})h_n^{*}(\boldsymbol{\beta})\delta(\boldsymbol{\alpha}-(\mathbf{d}_\x+\boldsymbol{\beta}))\dfd{\d}{\boldsymbol{\alpha}}\dfd{\d}{\boldsymbol{\beta}}\nonumber\\
&=\int_{\CN^\d}\lambda_{m}\lambda^{*}_{n}h_m(\boldsymbol{\beta}+\mathbf{d}_\x)h_n^{*}(\boldsymbol{\beta})\dfd{\d}{\boldsymbol{\beta}}.
\end{align}
Hence,
\begin{align} \LABEQ{app1}
\C_{U}(\x,\x\new)&=\sum_{m=1}^2\sum_{n=1}^2f_{mn}(\mathbf{d}_\x)
+\sum_{m=3}^4\sum_{n=3}^4f_{mn}(\mathbf{d}_\x).
\end{align}
The pseudo-covariance of $U(\x)$, $\tilde\C_{U}(\x,\x\new)=\E\left[U(\x)U(\x\new)\right]$ is derived in a similar way, and its calculation involves terms as
\begin{align}\LABEQ{ges}
g_{mn}(\mathbf{d}_\x)=\int_{\CN^\d}\lambda_{m}\lambda_{n}h_m(\boldsymbol{\beta}+\mathbf{d}_\x)h_n(\boldsymbol{\beta})\dfd{\d}{\boldsymbol{\beta}}.
\end{align}
One general example is to set the filters as parameterized exponentials, $h_i(\x)=v_{i}\exp(-(\x-\boldsymbol\mu_i)\her(\x-\boldsymbol\mu_i)/\gamma_i)$, so \EQ{efes} yields
\begin{align}\LABEQ{efescaso1}
&f_{mn}(\mathbf{d}_\x)=\nonumber\\
&\lambda_{m}\lambda_{n}^*v_{m}v_{n}\exp\left(-\frac{\left(\mathbf{d}_\x-\boldsymbol\mu_m+\boldsymbol\mu_n\right)\her\left(\mathbf{d}_\x-\boldsymbol\mu_m+\boldsymbol\mu_n\right)}{\gamma_m+\gamma_n}\right)\nonumber\\
&\cdot\left(\int_{\CN^\d}\exp\left(-\frac{(\gamma_m+\gamma_n)(\boldsymbol{\beta}-\hat{\boldsymbol\beta})\her(\boldsymbol{\beta}-\hat{\boldsymbol\beta})}{\gamma_m\gamma_n}\right)\dfd{\d}{\boldsymbol{\beta}}\right)\nonumber\\
&=\lambda_{m}\lambda_{n}^*v_{m}v_{n}\left(\frac{\pi\gamma_m\gamma_n}{\gamma_m+\gamma_n}\right)^{\d}\nonumber\\
&\cdot\exp\left(-\frac{\left(\mathbf{d}_\x-\boldsymbol\mu_m+\boldsymbol\mu_n\right)\her\left(\mathbf{d}_\x-\boldsymbol\mu_m+\boldsymbol\mu_n\right)}{\gamma_m+\gamma_n}\right)\nonumber\\
&=\lambda_{m}\lambda_{n}^*\bar{f}_{mn}(\mathbf{d}_\x),
\end{align}
where $\hat{\boldsymbol\beta}=(\boldsymbol\mu_n\gamma_m-(\mathbf{d}_\x-\boldsymbol\mu_m)\gamma_n)/(\gamma_m+\gamma_n)${, and 
\begin{align}\LABEQ{efemnbar}
\bar{f}_{mn}(\mathbf{d}_\x)&=v_{m}v_{n}\left(\frac{\pi\gamma_m\gamma_n}{\gamma_m+\gamma_n}\right)^{\d}\nonumber\\
&\cdot\exp\left(-\frac{\left(\mathbf{d}_\x-\boldsymbol\mu_m+\boldsymbol\mu_n\right)\her\left(\mathbf{d}_\x-\boldsymbol\mu_m+\boldsymbol\mu_n\right)}{\gamma_m+\gamma_n}\right).
\end{align}}
Analogous calculations yield 
\begin{align}\LABEQ{gescaso1}
&g_{mn}(\mathbf{d}_\x)=\lambda_{m}\lambda_{n}\bar{f}_{mn}(\mathbf{d}_\x).
\end{align}

If, as an example, $\boldsymbol\mu_m=\boldsymbol\mu_n=\matr{0}$ for all posible values of $m$ or $n$, after some simple mathematical manipulations,
\begin{align} \LABEQ{Kcaso1}
\C_{U}(\x,\x\new)&=\sum_{m=1}^{4}\bar{f}_{mm}(\mathbf{d}_\x),
\end{align}
where {$\bar{f}_{mm}(\mathbf{d}_\x)$ simplifies to}
\begin{align} \LABEQ{fmm}
\bar{f}_{mm}(\mathbf{d}_\x)=v_{m}^{2}\left(\frac{\pi\gamma_{m}}{2}\right)^{\d}\exp\left(-\frac{\mathbf{d}_\x\her\mathbf{d}_\x}{2\gamma_{m}}\right).
\end{align}
And the pseudo-covariance is
\begin{align} \LABEQ{pKcaso1}
\psc\C_{U}(\x,\x\new)&=\left(\bar{f}_{11}{(\mathbf{d}_\x)}+\bar{f}_{33}{(\mathbf{d}_\x)}\right)-\left(\bar{f}_{22}{(\mathbf{d}_\x)}+\bar{f}_{44}{(\mathbf{d}_\x)}\right)\nonumber\\
&+\j\left(2\bar{f}_{12}{(\mathbf{d}_\x)}+2\bar{f}_{34}{(\mathbf{d}_\x)}\right),
\end{align}
{where $\bar{f}_{mm}$ is given in \EQ{fmm}, and  $\bar{f}_{mn}$ now simplifies to
\begin{align}\LABEQ{efemnbar2}
\bar{f}_{mn}(\mathbf{d}_\x)&=v_{m}v_{n}\left(\frac{\pi\gamma_m\gamma_n}{\gamma_m+\gamma_n}\right)^{\d}\exp\left(-\frac{\mathbf{d}_\x\her\mathbf{d}_\x}{\gamma_m+\gamma_n}\right).
\end{align}}
Notice that in this example the covariance function $\k(\x,\x\new)=\C_{U}(\x,\x\new)$ is real{-valued} while the pseudo-covariance $\psc\k(\x,\x\new)=\psc\C_{U}(\x,\x\new)$ is complex-valued. This is due to the fact that $\K\jjrr=\K\rrjj\trs=\K\rrjj$ for the process generated with the filters in this example. 
The examples in \EQ{K1}-\EQ{pK1} are derived from \EQ{Kcaso1}-\EQ{pKcaso1} when 
$v_{1}=v_{3}=v_r$, $\gamma_{1}=\gamma_{3}=\gamma_r$, and $v_{2}=v_{4}=v_j$, $\gamma_{2}=\gamma_{4}=\gamma_j$.


In order to yield a complex covariance function, we need $\K\jjrr=\K\rrjj\trs\neq\K\rrjj$. An example arises when $\K\jjrr$ is skew-symmetric; $\K\jjrr=\K\rrjj\trs=-\K\rrjj$. In such a case the pseudo-covariance is real while \notaRBT{the} covariance is complex. In order to get a skew-symmetric $\K\jjrr$ there must be is a correlation between the real part and a displaced or translated imaginary part, with displacement given by $\boldsymbol{\mu} \in \CN, \boldsymbol{\mu}\neq\vect{0}$, while there is also a correlation between the real part and a displaced imaginary part when the displacement is given by $-\boldsymbol{\mu}$, and this correlation has the same value with the opposite sign. This is achieved with the following parameter values. For $h_1(\x)=h_3(\x)$ we set $\boldsymbol\mu_{1}=\boldsymbol\mu_{3}=\matr{0}$, $\gamma_{1}=\gamma_{3}=\gamma_r$ and $v_{1}=v_{3}=v\rr$. For $h_2(\x)$ we set $\boldsymbol\mu_{2}=\boldsymbol\mu$, $v_{2}=v\jj$ and $\gamma_{2}=\gamma_j$. And for $h_4(\x)$ we set $\boldsymbol\mu_{4}=-\boldsymbol\mu$, $v_{4}=-v\jj$ and $\gamma_{4}=\gamma_j$. {In this case, the covariance and pseudo-covariance yield}
\begin{align}
\C_{U}(\x,\x\new)&=2\bar{f}_{11}{(\mathbf{d}_\x)}+2\bar{f}_{22}{(\mathbf{d}_\x)}-\j2\left(\bar{f}_{12}{(\mathbf{d}_\x)}+\bar{f}_{34}{(\mathbf{d}_\x)}\right), \LABEQ{Kcaso2}\\
{\psc\C_{U}(\x,\x\new)}&={2\bar{f}_{11}(\mathbf{d}_\x)-2\bar{f}_{22}(\mathbf{d}_\x),}\LABEQ{pKcaso2}
\end{align} 
{where $\bar{f}_{mn}$ is given in \EQ{efemnbar}}.
The example in \EQ{kerfil}-\EQ{pkerfil} is derived from \EQ{Kcaso2}-\EQ{pKcaso2} when the inputs are complex-valued scalars $x\in\CC$, the displacement is also \notaRBT{a} complex-valued scalar $\mu\in\CC$, and $\gamma_{r}=\gamma_{j}=\gamma$.

\subsection{Gradient Descent of the Marginal Likelihood}\LABAPEN{Ap2} 
\notaRBT{The log marginal likelihood $L(\boldsymbol{\theta})$ in (\ref{eq:CMML}) is a function of a complex-valued Hermitian matrix $\C(\boldsymbol{\theta})$. Therefore, for its maximization we must seek generalized complex-valued matrix derivatives \cite{Hjorungnes08,Hjorungnes07}. We start by defining the following function }
 \begin{equation} \label{eq:CMML2}
g(\aug{\hat\C},\aug{\hat\C}^{*})=
-\frac{1}{2}\aug{\yv}\her\aug{{\hat \C}}\inv\aug{\yv}-\frac{1}{2}\log\det\aug{{\hat \C}}
\end{equation} 
where $\aug{\hat\C}$ is a matrix with independent components, i.e., not Hermitian. \notaRBT{The unpatterned matrix input variables $\aug{\hat\C}$ and $\aug{\hat\C}^{*}$ should be treated as independent when finding complex-valued matrix derivatives of the function $g(\aug{\hat\C},\aug{\hat\C}^{*})$.
We can find the derivatives of $L(\boldsymbol{\theta})$ in (\ref{eq:CMML}) with respect to $\aug{\C}$ and $\aug{\C}^*$ as follows \cite{Hjorungnes07}
\begin{align} 
\frac{\partial {L}}{\partial \aug{\C}} &=\left [{{\frac{\partial {g(\aug{\hat\C},\aug{\hat\C}^{*})}}{\partial \aug{\hat\C}}} }+{\left({\frac{\partial {g(\aug{\hat\C},\aug{\hat\C}^{*})}}{\partial \aug{\hat\C}^*}} \right)^\top}\right ]_{\aug{\hat\C}=\aug{\C}(\boldsymbol{\theta})}
,\label{eq:derivad1}
\end{align}
and
\begin{align} 
\frac{\partial {L}}{\partial \aug{\C}^*} &=\left [{{\frac{\partial {g(\aug{\hat\C},\aug{\hat\C}^{*})}}{\partial \aug{\hat\C}^*}} }+{\left({\frac{\partial {g(\aug{\hat\C},\aug{\hat\C})}}{\partial \aug{\hat\C}}} \right)^\top}\right ]_{\aug{\hat\C}=\aug{\C}(\boldsymbol{\theta})}
.\label{eq:derivad2}
\end{align}
Here the problem simplifies since $\partial{g(\aug{\hat\C},\aug{\hat\C}^{*})}/{\partial \aug{\hat\C}}^{*}=\vect{0}$. 
The derivative of $L(\boldsymbol{\theta})$ in (\ref{eq:CMML}) with respect to the hyperparameters is found by using the chain rule}
\begin{align} 
\frac{\partial {L}}{\partial \theta_i} &\notaRBT{=\mathrm{Tr}\left(\left(\frac{\partial {L}}{\partial \aug{\C}}\right)^\top\frac{\partial {\aug{\C}}}{\partial \theta_i}+\left(\frac{\partial {L}}{\partial \aug{\C}^*}\right)^\top\frac{\partial {\aug{\C}^*}}{\partial \theta_i}\right)}\nonumber\\
 &=\hspace{-0.05cm}2{\mathrm{Tr}\left(\hspace{-0.1cm}\left(\left.{\frac{\partial {g(\aug{\hat\C},\aug{\hat\C}^{*})}}{\partial \aug{\hat\C}}}\right|_{\aug{\hat\C}=\aug{\C}(\boldsymbol{\theta})} \right)^{\hspace{-0.08cm}\top}\hspace{-0.05cm}{\frac{\partial {\aug{\C}}}{\partial \theta_i}}\right)}.\label{eq:derivad}
\end{align}
The derivative of the first term of $g(\aug{\hat\C},\aug{\hat\C}^{*})$ with respect to $\hat\C$ yields
\begin{align} \label{eq:1term}
\frac{\partial {}}{\partial \aug{\hat\C}}\left(-\frac{1}{2}\aug{\yv}\her\aug{\hat\C}\inv\aug{\yv}\right)
&=\frac{1}{2}(\aug{\hat\C}^{\top})^{-1}\left(\aug{\yv}\,\aug{\yv}\her\right)^\top(\aug{\hat\C}^{\top})^{-1}.
\end{align}
The derivative of the second term of $g(\aug{\hat\C},\aug{\hat\C}^{*})$ with respect to $\hat\C$ yields
\begin{equation} \label{eq:2term}
\frac{\partial {}}{\partial \aug{\hat\C}}\left(-\frac{1}{2}\log\det\aug{\hat\C}\right)=-\frac{1}{2}(\aug{\hat\C}^{\top})^{-1}.
\end{equation}
Substitution of (\ref{eq:1term}) and (\ref{eq:2term}) in (\ref{eq:derivad}) yield (\ref{eq:gradientefinal}).
\ifCLASSOPTIONcaptionsoff
  \newpage
\fi



%

\bibliographystyle{IEEEtran}
\bibliography{IEEEabrv,CGPR,murilloGP,SSCDMA}

%
%

%








\end{document}